\documentclass[letterpaper, paper,11pt]{AAS}		

\usepackage{bm}
\usepackage{amsmath, amssymb}
\usepackage{amsfonts} 
\usepackage{bm}
\usepackage{mathtools}
\usepackage{siunitx}
\usepackage{subcaption}
\usepackage[colorlinks=true, pdfstartview=FitV, linkcolor=black, citecolor= black, urlcolor= black]{hyperref}
\usepackage{overcite}
\usepackage{footnpag}			
\usepackage{makecell}
\usepackage{algorithm}
\usepackage{algorithmic}

\usepackage{float}
\usepackage{graphicx}

\PaperNumber{25-868}

\begin{document}
\title{OPTIMAL THRUSTER CONFIGURATION FOR 6-DOF CONTROL OF A SMALL SATELLITE}

\author{Suguru Sato\thanks{Ph.D. Student, Mechanical and Aerospace Engineering, The University of Texas at Arlington, Arlington, TX 76019},  
Jinaykumar Patel\thanks{Ph.D. Student, Mechanical and Aerospace Engineering, The University of Texas at Arlington, Arlington, TX 76019},
\ and Kamesh Subbarao\thanks{Professor, Mechanical and Aerospace Engineering, The University of Texas at Arlington, Arlington, TX 76019}
}

\maketitle{} 		

\begin{abstract}
With the growing deployment of small satellites (such as CubeSats, Nanosats, Picosats, and Femtosats) in Low Earth Orbit (LEO) for targeted applications like imaging, communication, data storage, and rendezvous-docking mission, there is increasing attention on orbit maintenance and attitude control. A common approach for active orbit control involves the use of multiple thrusters, which, when properly arranged, can also generate the required torque for attitude control. Starting from a 24-thruster configuration, this paper presents a set of thruster configurations (referred to as a viable configuration group) that enable full six degrees of freedom (6-DOF) control. Further, configuration group that requires minimum total thrust to achieve 6-DOF commands are found among the viable configuration group. One configuration from each of these groups is further evaluated for its attitude control performance through a representative rendezvous-docking mission, demonstrating that even with a reduced thruster count, sufficient maneuverability can be achieved. 
\end{abstract}

\section{Introduction}
The successful capture of a tumbling object in orbit using an autonomous space vehicle has become an important research topic due to the continuous increase of orbit activities~\cite{nakasuka,subbaraoWelsh,nishida,romano}. Typical applications include collecting and removing space debris, servicing a malfunctioning satellite, refueling a powerless satellite, or installing an improved technology.~\cite{nasaOSAM} 

To gather position and state information for closed-loop control to navigate the capture vehicle toward the target, Lichter and Dubowsky~\cite{licheterDubowsky} developed an approach that uses 3-D vision sensors during an observation phase that provides input to a Kalman filter, which extracts the full dynamic state and inertial parameters of the target. Other approaches, particularly in a cooperative framework for space rendezvous, similarly utilized active vision~\cite{tsiotras}. 

For a small satellite, it is essential to have the capability of utilizing the full range of motion (6-DOF) while serving as a capture vessel. For typical satellites, it is common to have a combination of active (thrusters, control moment gyros) and passive (reaction wheels) modalities for 6-DOF maneuvering. 

Theoretical results from multi-jet spacecraft control studies indicate that to achieve $n$-DOF, a minimum of $n+1$ unidirectional actuators (such as thrusters) are required to ensure complete controllability in the absence of redundancy~\cite{crawford1969, crawford1969operation}. This principle establishes that for a 6-DOF system, at least seven unidirectional thrusters are necessary to span the full activity space, ensuring the ability to generate arbitrary forces and torques in all directions. In this paper, we verify this theoretical result using a heuristic approach, evaluating whether a configuration with 7 thrusters can consistently achieve full 6-DOF control for a cubic satellite through an iterative analysis of thruster placements and their resultant force and torque capabilities. In addition, we investigate the optimal number of thrusters in terms of the balance between the number of thrusters, controllability, and thrust efficiency.

In this paper, we restrict attention to only the use of thrusters for accomplishing both translational and attitude control. Since the choice of the number and placement of these thrusters is a design choice - balancing the weight, overall complexity~\cite{yoshimura}, control allocation~\cite{harkegard}, and control authority; the paper provides a method to derive an optimal number of thrusters (quantity and placement) for a small satellite (a cubic satellite in particular). 

\section{Problem Statement}
While placing a large number of thrusters on a satellite can easily ensure full 6-DOF motion, this approach is not efficient, as many thrusters may end up being overly redundant. This raises an important and nontrivial question: \textit{What is the minimum number of thrusters required to enable both translational and rotational motion in any desired direction?} Another key question would be: \textit{Is the configuration with the minimum number of thrusters optimal?} This paper addresses these questions by developing an algorithm to optimize thruster count starting from an arbitrary initial configuration.

As the initial configuration, a satellite is equipped with 24 thrusters. These thrusters are symmetrically arranged with respect to the satellite’s three body planes (X-Y, X-Z, Y-Z), with four thrusters placed on each face of the cube. These symmetrically installed 24 thrusters guarantee 6-DOF control of the initial configuration.

The configurations presented as the final solutions in this paper represent the one that achieves the lowest thruster count, and the one that ensures the optimal balance between the number of thrusters, controllability, and thrust efficiency. In the initial 24-thruster setup, each of the satellite’s vertices is equipped with three thrusters. For the sake of mathematical simplicity, the thrust vectors are assumed to lie perpendicular to the satellite's faces without gimbal capability. With this initial configuration established, the paper proceeds to summarize the governing equations of motion.

To verify the practicality of these solutions, a representative configuration from each thruster count is evaluated in a simulated rendezvous-docking mission. In this mission, the chaser satellite must precisely approach a tumbling target, matching both its relative position and attitude while accounting for orbital dynamics and maximum thrust. This evaluation compares key performance metrics such as control accuracy and thrust profiles across configurations, highlighting how reduced thruster sets perform in real-world proximity operations, thereby bridging theoretical optimality with operational viability.

\section{Equations of Motion}
For a satellite in LEO, the sum of all the forces acting on it can be written as:
\begin{equation}
    {\bm f} = {\bm f}_g + {\bm f}_c + {\bm f}_p
\end{equation}
where \({\bm f}_g \in \mathbb{R}^{3 \times 1}\) is the force due to gravity, \({\bm f}_c\) is the control force, and \({\bm f}_p\) is the force due to perturbations (unmodeled bounded disturbances). Assuming \({\bm f}_p = \bm 0_{3 \times 1}\), and a Keplerian orbit for the satellite, the translational dynamics can be described as follows.
\begin{equation}
    \ddot{\bm r}_s = - \frac{\mu}{r_s^3}{\bm r}_s + \frac{1}{m}{\bm f}_c
    \label{eq:translational_dynamics}
\end{equation}
where \({\bm r}_s \in \mathbb{R}^{3 \times 1}\) is the position vector of the satellite with respect to the center of the Earth, \(r_s = \|{\bm r}_s\|\), and \(m\) is the mass of the satellite. 

The rotational motion of the satellite can be written as follows.
\begin{equation}
    {\bm I}_s\dot{\bm \omega}_s + {\bm \omega}_s \times {\bm I}_s{\bm \omega}_s = {\bm \tau}_c + {\bm \tau}_p
\end{equation}
where \({\bm I}_s \in \mathbb{R}^{3 \times 3}\) and \({\bm \omega}_s \in \mathbb{R}^{3 \times 1}\) are the moment of inertia and the angular rate of the satellite respectively, \({\bm \tau}_c \in \mathbb{R}^{3 \times 1}\) is the control torque, and \({\bm \tau}_p\) is the moment due to perturbations.

\begin{figure}[htb]
    \centering
    \begin{subfigure}{0.45\textwidth}
    \centering
        \includegraphics[width=0.5\linewidth]{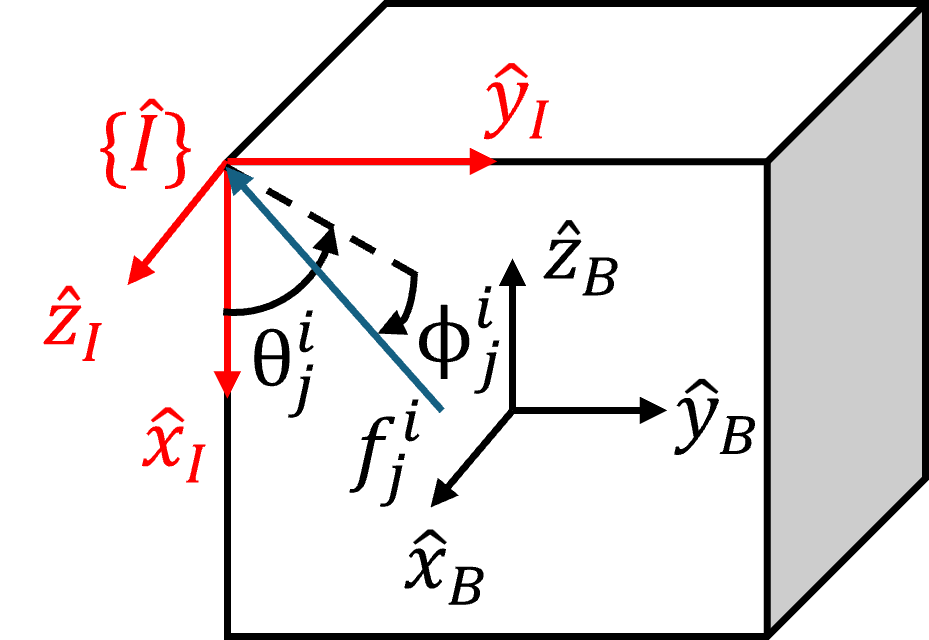} 
        \caption{Satellite's coordinate systems and rotation angles}
        \label{fig:Figure1}
    \end{subfigure}
    \begin{subfigure}{0.45\textwidth}
    \centering
        \includegraphics[width=0.6\textwidth]{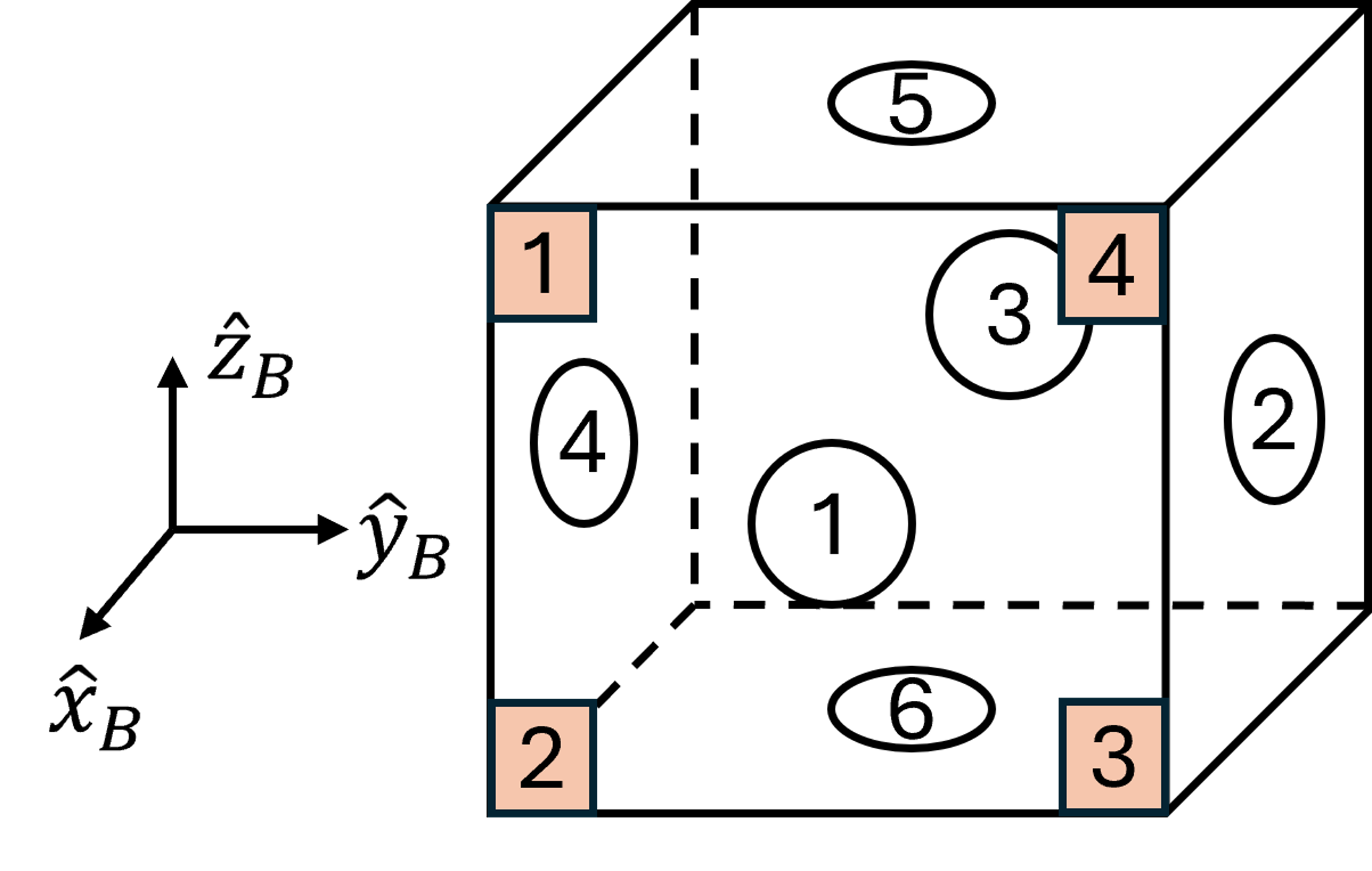}
        \caption{Satellite's face number and corner number allocation}
        \label{fig:number_allocation}
    \end{subfigure}
    \caption{Satellite's geometric notations}
    \label{fig:example_configurations}
\end{figure}

With the given equations of motion, it is of interest to find a model for the control force and moments. We assuming \({\bm f}_p = \bm 0_{3 \times 1}\) and \( {\bm \tau}_p = \bm 0_{3 \times 1}\) for simplicity. As mentioned previously, there exist 24 thrusters on the surface of the satellite for the initial configuration. On each face of the satellite, there are 4 thrusters placed at each of the corners. The orientation of these thrusters can be described by the face-azimuth angle \(\theta\) and the face-elevation angle \(\phi\) with respect to the local frame attached to each of the satellite's faces as shown in Figure \ref{fig:Figure1}. Then, the thrust force due to a particular thruster can be expressed as shown in Eq.~(\ref{eq:thruster_in_face}).

\begin{equation}
    {\bm f}_{j,\hat{\bm I}}^i = 
    \begin{Bmatrix}
        -\cos{\phi_j^i}\cos{\theta_j^i} \\
        -\cos{\phi_j^i}\sin{\theta_j^i} \\
        -\sin{\phi_j^i}
    \end{Bmatrix}
    f_j^i
    = {\bm d}(\phi_j^i, \theta_j^i) f_j^i
    \label{eq:thruster_in_face}
\end{equation}
where \(f_j^i\) represents force of a thruster on \(j\) th corner on \(i\) th face (\(i\) = 1, \ldots, 6, \(j\) = 1, \dots, 4), and \(\hat{\bm I}\) represents satellite's face frame. Additionally, the thruster ID is defined as \((i-1)\cdot4 + j\) (e.g. a thruster located at corner 4 on face 3 is assigned a thruster ID of 12. Conversely, the 15th thruster corresponds to face 4 and corner 3, respectively.). By denoting \(\gamma^i\) and \(\delta^i\) as rotation angles from the face fixed frame, \(\hat{\bm I}\), to the satellite's body frame, \(\hat{\bm B}\), a thrust can be transformed into the satellite's body frame as follows by applying transformation matrix \({\bm C}(\gamma^i, \delta^i)\).
\begin{equation}
    {\bm f}_{j,\hat{\bm B}}^i = 
    \begin{bmatrix}
        {\bm C}(\gamma^i, \delta^i)
    \end{bmatrix}
    {\bm f}_{j,\hat{\bm I}}^i
    =
    \begin{bmatrix}
        {\bm C}(\gamma^i, \delta^i)
    \end{bmatrix}
    {\bm d(\phi_j^i, \theta_j^i)} f_j^i
    \label{eq:thrust_in_body}
\end{equation}

By substituting Eq.~(\ref{eq:thrust_in_body}) into Eq.~(\ref{eq:translational_dynamics}), the expression of the translational dynamics can be rewritten as follows.
\begin{equation}
    \ddot{\bm r}_s + \frac{\mu}{r_s^3}{\bm r}_s = \frac{1}{m}{\bm f}_c = \frac{1}{m} \sum_{i=1}^6 \sum_{j=1}^{4} 
    \left(
    \begin{bmatrix}
        {\bm C}(\gamma^i, \delta^i)
    \end{bmatrix} {\bm d(\phi_j^i, \theta_j^i)} f_j^i
    \right) 
\end{equation}

Now, by denoting \( {\bm C}(\gamma^i, \delta^i) \) as \( {\bm C}_i \) and \(\bm d(\phi_j^i, \theta_j^i) \) as \({\bm d}_j^i\) for short, the full 24-thruster translational dynamics is led to the following.
\begin{equation}
    \ddot{\bm r}_s + \frac{\mu}{r_s^3}{\bm r}_s = \frac{1}{m}{\bm f}_c 
    = \frac{1}{m}
    \begin{Bmatrix}
        f_x \\
        f_y \\
        f_z
    \end{Bmatrix}
    =\frac{1}{m}
    \begin{bmatrix}
        [{\bm C}_1]{\bm d}_1^1, \ldots, [{\bm C}_6]{\bm d}_4^6
    \end{bmatrix}_{3 \times 24}
    \begin{Bmatrix}
        f_1^1 \\
        \vdots \\
        f_4^6 \\
    \end{Bmatrix}_{24 \times 1}
    \label{eq:force_allocation}
\end{equation}

Thus, the force allocation problem is clearly evident from Eq.~(\ref{eq:force_allocation}). Now, let us develop the rotational dynamics of the satellite. Let the position vector of each thruster with respect to the center of mass of the satellite in the satellite body frame be
\begin{equation}
    {\bm r}_j^i = 
    \begin{bmatrix}
        r_{j,x}^i \;, & r_{j,y}^i \;, & r_{j,z}^i
    \end{bmatrix}^T~.
\end{equation}

The moment generated by a thruster is calculated by
\begin{equation}
    {\bm \tau}_j^i = {\bm r}_j^i \times {\bm f}_{j,\hat{\bm B}}^i~.
    \label{eq:9}
\end{equation}
From Eq. (\ref{eq:thrust_in_body}), above equation can be rewritten as
\begin{equation}
    {\bm \tau}_j^i = {\bm r}_j^i \times 
    \begin{bmatrix}
        {\bm C}_i
    \end{bmatrix}
    {\bm d}_j^i f_j^i~.
\end{equation}

By denoting the skew-symmetric matrix of \({\bm r}_j^i\) as \({\bm B}_j^i\), the full 24-thruster rotational dynamics can be rewritten as
\begin{equation}
    {\bm I}_s\dot{\bm \omega}_s + {\bm \omega}_s \times {\bm I}_s{\bm \omega}_s = {\bm \tau}
    =
    \begin{Bmatrix}
        \tau_x \\ \tau_y \\ \tau_z
    \end{Bmatrix}
    = 
    \begin{bmatrix}
        [{\bm B}_1^1][{\bm C}_1]{\bm d}_1^1, \ldots, [{\bm B}_4^6][{\bm C}_6]{\bm d}_4^6
    \end{bmatrix}_{3 \times 24}
    \begin{Bmatrix}
        f_1^1 \\
        \vdots \\
        f_4^6 \\
    \end{Bmatrix}_{24 \times 1}~,
    \label{eq:rotational_dynamics}
\end{equation}

Finally, the forces and moments required for the full translational and rotational motion of the satellite are given by
\begin{equation}
    \begin{Bmatrix}
        f_x \\ f_y \\ f_z \\ \tau_x \\ \tau_y \\ \tau_z
    \end{Bmatrix}
    =
    \begin{bmatrix}
         [{\bm C}_1]{\bm d}_1^1 & \ldots &  [{\bm C}_1]{\bm d}_4^1 & \dots &  [{\bm C}_6]{\bm d}_1^6  & \ldots &  [{\bm C}_6]{\bm d}_4^6 \\
        [{\bm B}_1^1][{\bm C}_1]{\bm d}_1^1 & \ldots & [{\bm B}_4^1][{\bm C}_1]{\bm d}_4^1  & \ldots & [{\bm B}_1^6][{\bm C}_6]{\bm d}_1^6 & \ldots & [{\bm B}_4^6][{\bm C}_6]{\bm d}_4^6
    \end{bmatrix}
    \begin{Bmatrix}
        f_1^1 \\
        \vdots \\
        f_4^1 \\
        \vdots \\
        f_1^6 \\
        \vdots \\
        f_4^6 \\
    \end{Bmatrix}
    \label{eq:combined_eq}
\end{equation}

Let \(\mathcal{H} = 
    \begin{bmatrix}
         [{\bm C}_1]{\bm d}_1^1 & \ldots &  [{\bm C}_1]{\bm d}_4^1 & \dots &  [{\bm C}_6]{\bm d}_1^6  & \ldots &  [{\bm C}_6]{\bm d}_4^6 \\
        [{\bm B}_1^1][{\bm C}_1]{\bm d}_1^1 & \ldots & [{\bm B}_4^1][{\bm C}_1]{\bm d}_4^1  & \ldots & [{\bm B}_1^6][{\bm C}_6]{\bm d}_1^6 & \ldots & [{\bm B}_4^6][{\bm C}_6]{\bm d}_4^6
    \end{bmatrix}
\) and \(\mathcal{H} \in \mathbb{R}^{6 \times 24}\). Thus, all the orientation and position parameters of the 24 thrusters are within the \(\mathcal{H}\). As it can be seen from Eq. (\ref{eq:combined_eq}), there are multiple possible solutions for the translational and rotational parameters that achieve commanded force and torque given in the LHS. Since the rank of a matrix indicates the number of linearly independent output directions it can produce from its input, \(\mathcal{H}\) needs to have rank 6 for full 6-DOF solutions to be possible. Further, several combinations of thrusters can be employed to achieve this. Note that the thrusters are only assumed to produce a unidirectional thrust; hence, the solutions sought always have positive values. Thus, this problem is solved by seeking a \textit{non-negative least squares solution}.

\section{Solution Methodology}
With the \(\mathcal{\bm H} \in \mathbb{R}^{6 \times 24}\) matrix derived in the prior section, the thruster count optimization can be performed. As a candidate approach, a heuristic algorithm is adopted.

\begin{enumerate}
    \item \textbf{Combinations of thrusters}: Generate all possible combinations of thruster indices out of the 24 thrusters (e.g. There are ${}^{24}C_N$ combinations for an N-thruster configuration).
    \item \textbf{Reconstruction of \(\mathcal{\bm H}\)}: For each of the combinations, reconstruct the control allocation matrix, \(\mathcal{\bm H}\), with the entries corresponding to the active thrusters. Let us call this control allocation matrix of \(\ell\)-th combination as \(\mathcal{\bm H}_{\ell}  \in \mathbb{R}^{6 \times N}\). 
    \item \textbf{Rank of \(\mathcal{\bm H}_{\ell}\)}: To reduce the number of candidate solutions, the rank of the \(\mathcal{\bm H}_{\ell}\) is checked. If its rank is less than 6, it is thrown out, as it does not support full 6-DOF.
    \item \textbf{Capability for pure forces and moments}: For valid \(\mathcal{\bm H}_{\ell}\) matrices, we check to see if they can produce unit forces and moments without a thruster needing to fire negatively, as this is against the assumption of unidirectional thruster. This is done for all forces and moments, positive and negative, about the satellite's X, Y, and Z axes since a full range of motion is required. To perform this step, we employ the nonnegative least squares algorithm. For example, for the \(\ell\)-th combination, nonnegative least squares is performed as follows to achieve a unit positive force in the satellite's X axis:
    \begin{equation}
        {\bm f}_{\ell,f_x^+} = \arg \underset{f_{\ell}\geq 0}{\min} \| \mathcal{\bm H}_{\ell}{\bm f}_{\ell} - {\bm u}_{c,f_x^+} \|_2^2
        \label{eq:13}
    \end{equation}
    where \({\bm f}_\ell \in \mathbb{R}^{N \times 1}\) is a vector of the active thruster forces to be solved, \({\bm f}_{\ell,f_x^+}  \in \mathbb{R}^{N \times 1} \) is a vector of the optimized active thruster forces to generate unit positive force in the satellite's X-body direction, and \({\bm u}_{c,f_x^+}  \in \mathbb{R}^{6 \times 1}\) is a unit command force vector in the satellite's X-body direction (i.e. \({\bm u}_{c,f_x^+} = [1, 0, 0, 0, 0, 0]^T\) ). This nonnegative least squares algorithm is run for each of the 12 sets of positive and negative force and moment commands to obtain the required thrusts, \({\bm f}_{\ell,f_x^+}, {\bm f}_{\ell,f_y^+},\ldots, {\bm f}_{\ell,\tau_y^-}, {\bm f}_{\ell,\tau_z^-}\). 
    \item \textbf{Check residual}: Since the results from nonnegative least squares are the results of optimization, their residuals need to be checked to validate their effectiveness. For example, the residual about \( {\bm f}_{\ell,f_x^+} \) is checked as follows:
    \begin{equation}
        \| {\bm r}_{\ell,f_x^+} \|_2^2= \| {\bm u}_{c,f_x^+} - \mathcal{\bm H}_\ell {\bm f}_{\ell,f_x^+} \|_2^2 
    \end{equation}

    After the residual for each of the optimization results (i.e. \({\bm f}_{\ell,f_x^+}, \ldots, {\bm f}_{\ell,\tau_z^-}\)) is calculated, if any of the residual exceeds a certain threshold $\epsilon_{NNLS}$, the combination is thrown out as it fails to execute the assigned commands precisely. Configurations passed this step are categorized as \textit{viable configuration} as they can achieve 6-DOF commands precisely.
    \item \textbf{Optimal configuration}: Although multiple viable configurations are found for a configuration with a fixed number of thrusters, thrusts that achieve the unit commands (i.e. results from Eq. (\ref{eq:13})), differ from one configuration to other as some configurations require more effort on thrusters than other configuraitons do. In this step, the configurations that require the least total thrust to achieve the unit commands are identified. The total thrust for achieving the unit commands is calculated as follows:
    \begin{equation}
        f_{\ell,T} = \sum_{p=1}^N\sum_{q=1}^{12} {\bm F}_{\ell,C}[p, q]
    \end{equation}
    where \({\bm F}_{\ell,C}\) is a matrix that combines the results from step 4 for \(\ell\)-th combination, and 
    \begin{equation}
        {\bm F}_{\ell,C} = \begin{bmatrix}
            {\bm f}_{\ell,f_x^+}, {\bm f}_{\ell,f_y^+},\ldots, {\bm f}_{\ell,\tau_y^-}, {\bm f}_{\ell,\tau_z^-}
        \end{bmatrix}
    \end{equation}
    and \({\bm F}_{\ell,C} \in \mathbb{R}^{N \times 12}\). Rows and columns of \({\bm F}_{\ell,C}\) are represented as \(p\) and \(q\), respectively.
    Then, the combinations that result in the smallest \(f_{\ell,T}\) will be categorized as \textit{optimal configurations}. Let us denote this smallest value of \(f_{\ell,T}\) for the \(N\)-thruster setting as \(f_{N,min}\).
    \item \textbf{Repeat the process}: Starting from the satellite equipped with 24 thrusters, this process is repeated until the number of thrusters is reduced until no viable configuration is found. 
\end{enumerate}

The complete optimization problem for determining the most efficient thruster configuration, minimizing total thrust effort, can be formally stated as follows, with an inner optimization that identifies the minimum number of thrusters required to achieve full 6-DOF controllability.
\begin{equation} \label{eq:your_main_label}
\begin{aligned}
\min_{\mathcal{S}_\ell} \quad & f_{\ell,T} = \sum_{i=1}^{N}\sum_{j=1}^{12} \bm{F}_{\ell,C}[i,j]\\
\text{s.t.} \quad & \\
& \mathcal{S}_\ell \subseteq \{1,\dots,24\}, \quad |\mathcal{S}_\ell| = N \\
&\mathcal{\bm W}
  \;=\;
  \bigl\{f_x^+, f_y^+, f_z^+\; \tau_x^+, \tau_y^+, \tau_z^+\; f_x^-, f_y^-, f_z^-\; \tau_x^-, \tau_y^-, \tau_z^-\bigr\}\\
& \mathcal{H}_\ell = \begin{bmatrix} 
    [\mathbf{C}_1]\mathbf{d}_1^1 & \cdots & [\mathbf{C}_6]\mathbf{d}_4^6 \\ 
    [\mathbf{B}_1^1][\mathbf{C}_1]\mathbf{d}_1^1 & \cdots & [\mathbf{B}_4^6][\mathbf{C}_6]\mathbf{d}_4^6 
    \end{bmatrix}_{\mathcal{S}_\ell} \in \mathbb{R}^{6 \times N} \\
& rank(\mathcal{\bm H}_\ell) = 6 \\
& {\bm f}_{\ell,m} \geq 0 \quad \forall m \in \mathcal{\bm W} \\
& \mathcal{\bm H}_\ell f_{\ell,m} = u_{c,m} \quad \forall m \in \mathcal{\bm W} \\
& \|{\bm u}_{c,m} - \mathcal{\bm H}_\ell {\bm f}_{\ell,m}\|_2^2 \leq \epsilon_{\mathrm{tol}} \quad \forall m \in \mathcal{\bm W}
\end{aligned}
\end{equation}

\section{Configuration Result}
The outcomes of the proposed approach are summarized in Table~\ref{tab:outcome}. Starting from the 24-thruster configuration, the lowest thruster count found was 7 thrusters, using \(\theta\) = 0\(^\circ\) and \(\phi\) = 90\(^\circ\) (that is, all possible thrusters are perpendicular to the faces to which they are attached). Although a variety of thruster orientations for 6-thruster configurations were tested, no viable configuration was found. For a cubic satellite with 7 thrusters, 48 viable configurations were found. A few of the viable 7-thruster configurations are shown below as examples in Fig.~\ref{fig:example_configurations}.
\\
\begin{figure}[htb]
    \centering
    \begin{subfigure}{0.3\textwidth}
        \includegraphics[trim=50 20 30 50,width=1.0\textwidth]{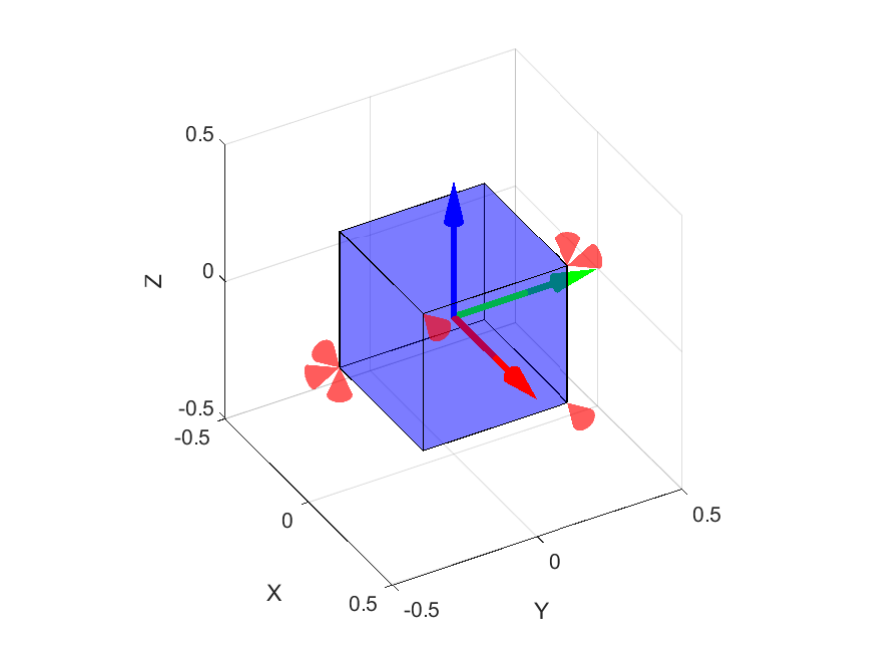}
        \caption{Configuration 1}
        \label{fig:config1}
    \end{subfigure}
    \begin{subfigure}{0.3\textwidth}
        \includegraphics[trim=50 20 30 50,width=1.0\textwidth]{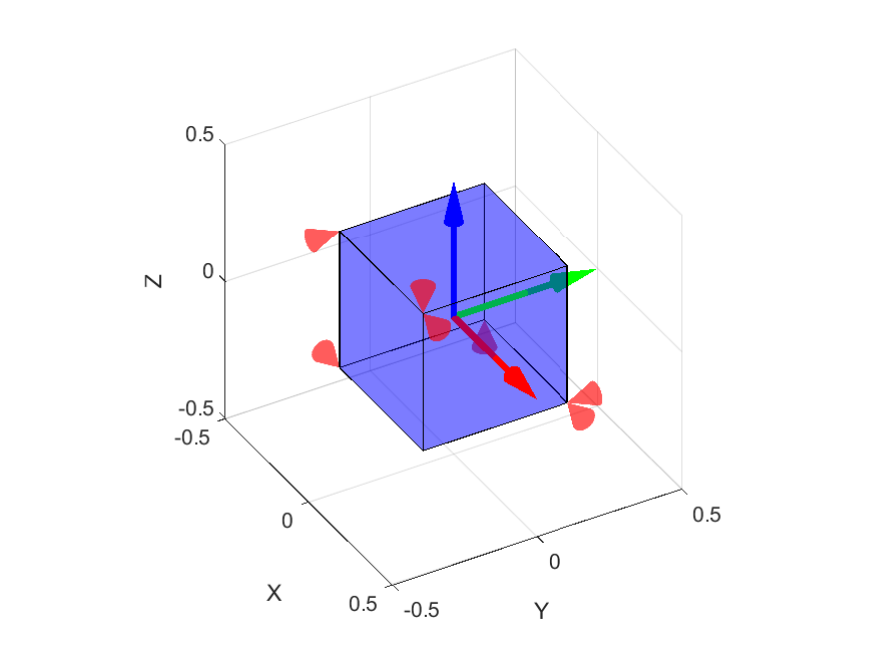}
        \caption{Configuration 2}
        \label{fig:config2}
    \end{subfigure}
    \begin{subfigure}{0.3\textwidth}
        \includegraphics[trim=50 20 30 50,width=1.0\textwidth]{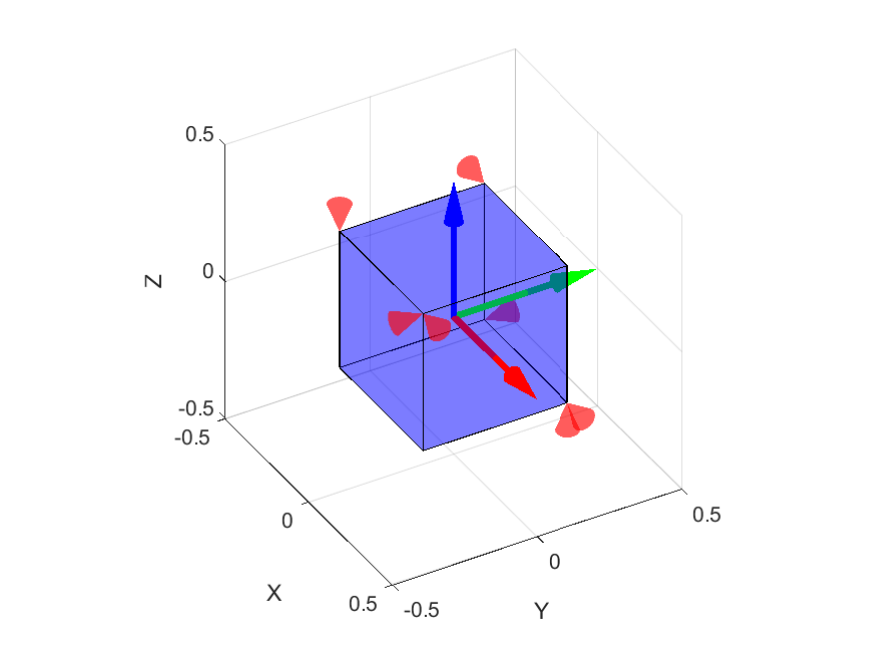}
        \caption{Configuration 3}
        \label{fig:config3}
    \end{subfigure}
    \caption{A few of 6-DOF cubic satellites, 7-thruster viable configurations}
    \label{fig:example_configurations}
\end{figure}

\begin{table}[htb]
\small
\centering
\caption{Summary of the configuration solutions}
\label{tab:outcome}
\begin{tabular}{ccccc}
\hline
\textbf{No. of Thrusters} & \textbf{ Combinations} & \textbf{Viable Configurations} & \textbf{Optimal Configurations} & $f_{N,\min}$ \textbf{(N)} \\
\hline
6  & 134596 & 0 & 0 & -- \\
7  & 346104 & 48 & 48 & 68 \\
8  & 735471 & 1536 & 47 & 38 \\
9  & 1307504 & 15040 & 256 & 36 \\
10 & 1961256 & 79572 & 368 & 34 \\
11 & 2496144 & 262128 & 160 & 32 \\
12 & 2704156 & 579864 & 24 & 30 \\
13 & 2496144 & 904272 & 279 & 30 \\
14 & 1961256 & 1034364 & 1605 & 30 \\
15 & 1307504 & 894400 & 5510 & 30 \\
16 & 735471 & 597294 & 12367 & 30 \\
17 & 346104 & 312432 & 17560 & 30 \\
18 & 134596 & 128912 & 18282 & 30 \\
19 & 42504 & 41904 & 12005 & 30 \\
20 & 10626 & 10596 & 10596 & 30 \\
21 & 2024 & 2024 & 1425 & 30 \\
22 & 276 & 276 & 248 & 30 \\
23 & 24 & 24 & 24 & 30 \\
24 & 1 & 1 & 1 & 30 \\
\hline
\end{tabular}
\end{table}

\begin{figure}[htb!]
    \centering
    \begin{subfigure}{0.3\textwidth}
        \includegraphics[trim=50 20 30 50,width=1.0\textwidth]{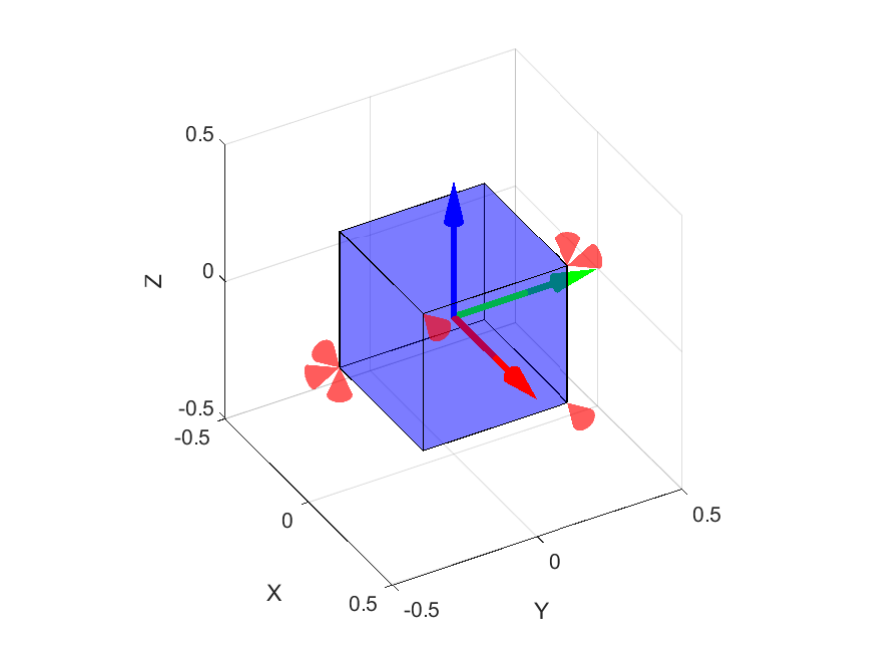}
        \caption{\(N=7\) optimal config.}
        \label{fig:optimalConfig7}
    \end{subfigure}
    \begin{subfigure}{0.3\textwidth}
        \includegraphics[trim=50 20 30 50,width=1.0\textwidth]{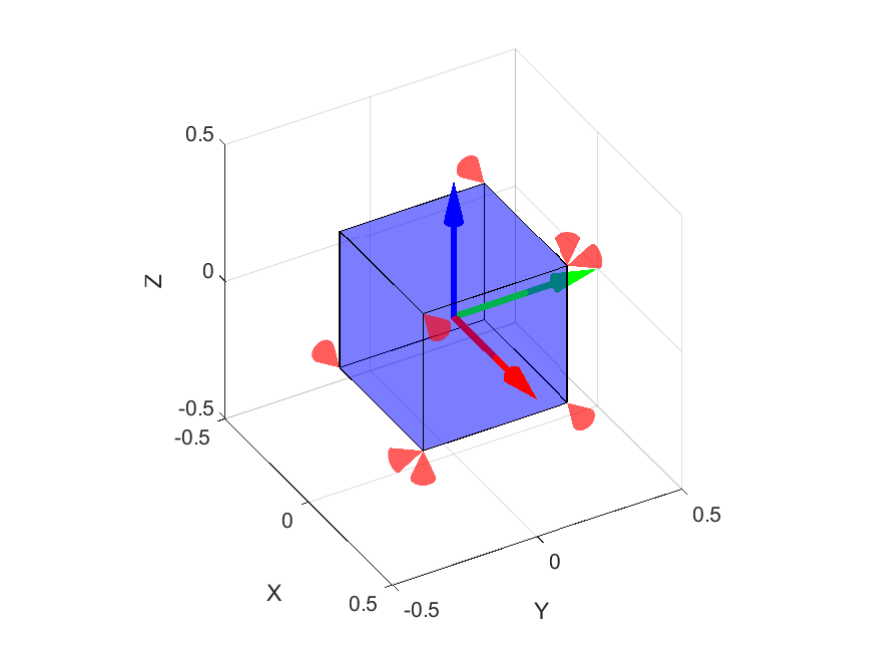}
        \caption{\(N=8\) optimal config.}
        \label{fig:optimalConfig8}
    \end{subfigure}
    \begin{subfigure}{0.3\textwidth}
        \includegraphics[trim=50 20 30 50,width=1.0\textwidth]{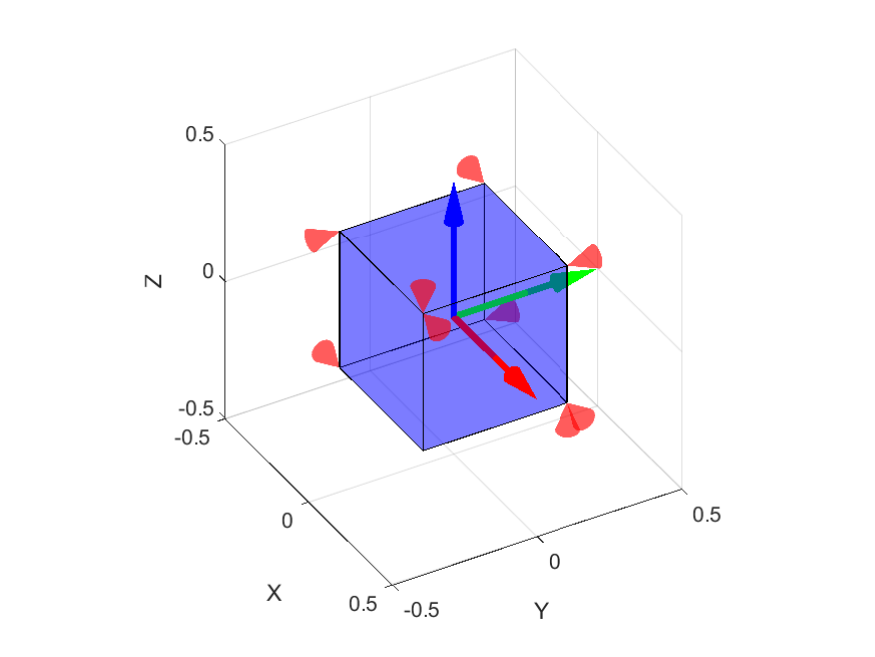}
        \caption{\(N=9\) optimal config.}
        \label{fig:optimalConfig9}
    \end{subfigure}
    \vskip\baselineskip
    \begin{subfigure}{0.3\textwidth}
        \includegraphics[trim=50 20 30 50,width=1.0\textwidth]{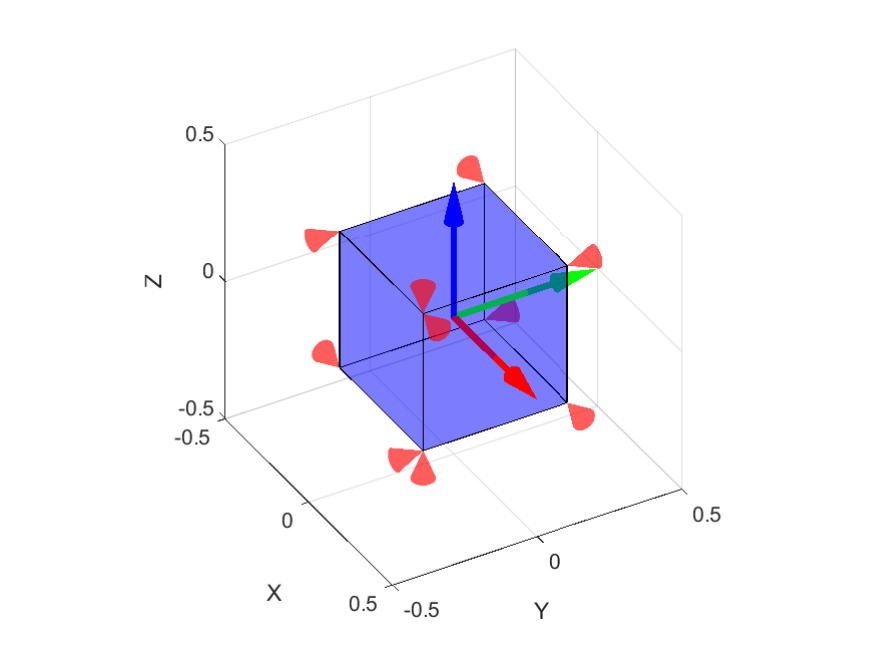}
        \caption{\(N=10\) optimal config.}
        \label{fig:optimalConfig10}
    \end{subfigure}
    \begin{subfigure}{0.3\textwidth}
        \includegraphics[trim=50 20 30 50,width=1.0\textwidth]{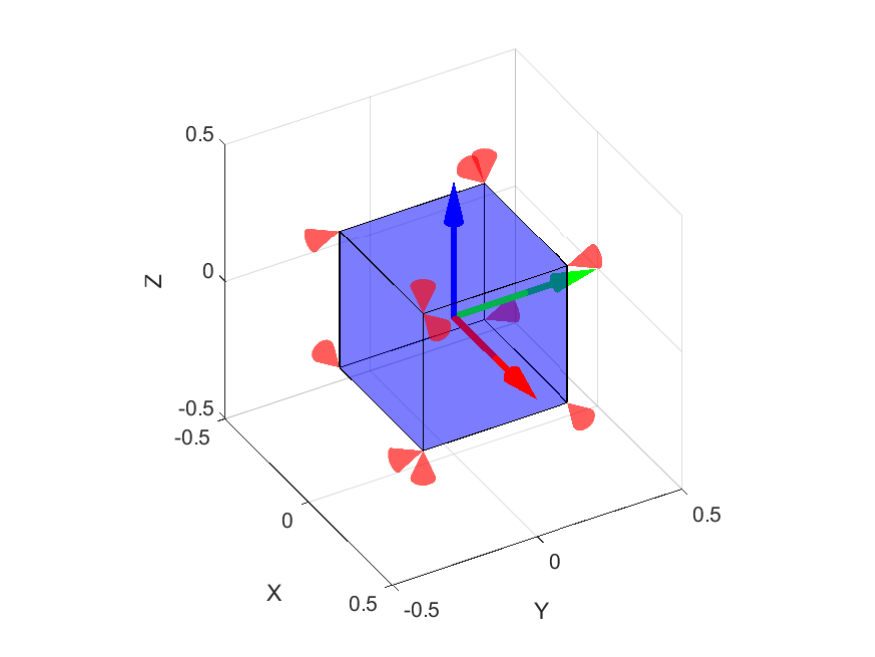}
        \caption{\(N=11\) optimal config.}
        \label{fig:optimalConfig11}
    \end{subfigure}
    \begin{subfigure}{0.3\textwidth}
        \includegraphics[trim=50 20 30 50,width=1.0\textwidth]{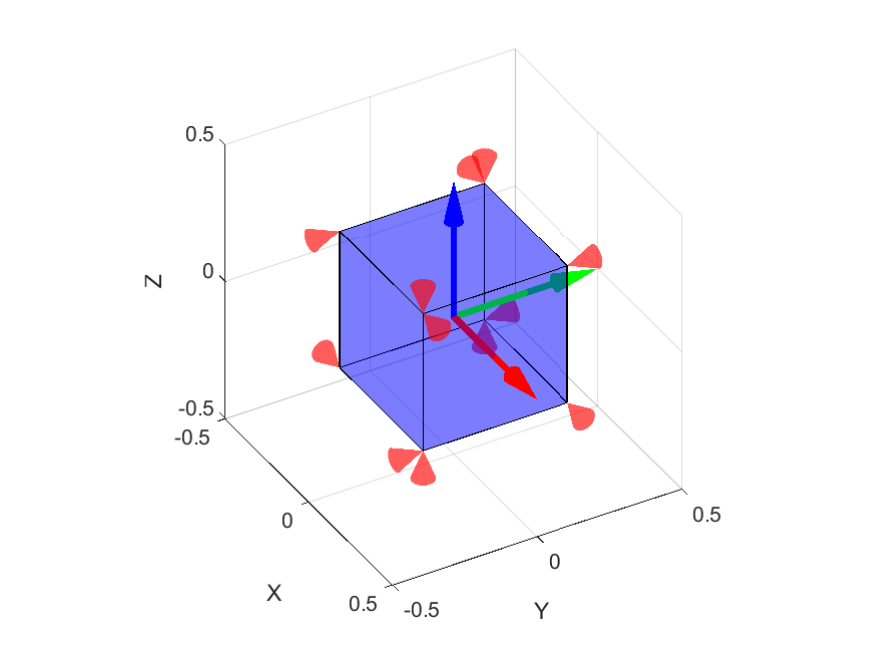}
        \caption{\(N=12\) optimal config.}
        \label{fig:optimalConfig12}
    \end{subfigure}
    \caption{One of the optimal configurations from \(N=7\) to \(12\) thruster count}
    \label{fig:optimal_configurations}
\end{figure}

The notable result from Table~\ref{tab:outcome} is the trend of \(f_{N,min}\). From \(N=7\) to \(12\), \(f_{N,min}\) decreases monotonically. However, at \(N=12\) and onward, it stays 30N. This result suggests that the least number of thrusters that support 6-DOF is 7 (which aligns with the theory); however, the optimal number of thrusters may be 12 instead of 7. Given this result, one of the optimal configurations from \(N = 7\) to \(12\) is showcased in Fig.~\ref{fig:optimal_configurations}. The thruster IDs of one of the optimal configurations from each thruster count are provided in Table \ref{tab:opt_thrusterID}. To further evaluate this result and observation, those optimal configurations given in Table \ref{tab:opt_thrusterID}, ranging from \(N=7\) to \(24\), are tested in the rendezvous-docking mission in the next section.

\begin{table}[htb!]
\small
\centering
\caption{Thruster ID of the optimal configurations}
\label{tab:opt_thrusterID}
\begin{tabular}{cccc}
\hline
\textbf{No. of Thrusters} & \textbf{Thruster ID}\\
\hline
7  & $[1,3,5,11,14,18,24]$ ~$\rightarrow$~ Fig.\ref{fig:optimalConfig7}\\
8  & $[1,3,5,9,11,15,18,21]$ ~$\rightarrow$~ Fig.\ref{fig:optimalConfig8} \\
9  & $[1,3,5,7,9,11,13,17,22]$ ~$\rightarrow$~ Fig.\ref{fig:optimalConfig9} \\
10 & $[1,3,5,7,9,11,13,15,17,21]$ ~$\rightarrow$~ Fig.\ref{fig:optimalConfig10} \\
11 & $[1,3,5,7,9,11,13,15,17,19,21]$ ~$\rightarrow$~ Fig.\ref{fig:optimalConfig11} \\
12 & $[1,3,5,7,9,11,13,15,17,19,21,23]$ ~$\rightarrow$~ Fig.\ref{fig:optimalConfig12} \\
13 & $[1,2,3,5,7,9,11,13,15,17,19,21,23]$ \\
14 & $[1,2,3,4,5,7,9,11,13,15,17,19,21,23]$ \\
15 & $[1,2,3,4,5,6,7,9,11,13,15,17,19,21,23]$ \\
16 & $[1,2,3,4,5,6,7,8,9,11,13,15,17,19,21,23]$ \\
17 & $[1,2,3,4,5,6,7,8,9,10,11,13,15,17,19,21,23]$ \\
18 & $[1,2,3,4,5,6,7,8,9,10,11,12,13,15,17,19,21,23]$ \\
19 & $[1,2,3,4,5,6,7,8,9,10,11,12,13,14,15,17,19,21,23]$ \\
20 & $[1,2,3,4,5,6,7,8,9,10,11,12,13,14,15,16,17,19,21,23]$ \\
21 & $[1,2,3,4,5,6,7,8,9,10,11,12,13,14,15,16,17,18,19,21,23]$ \\
22 & $[1,2,3,4,5,6,7,8,9,10,11,12,13,14,15,16,17,18,19,20,21,23]$ \\
23 & $[1,2,3,4,5,6,7,8,9,10,11,12,13,14,15,16,17,18,19,20,21,22,23]$ \\
24 & $[1,2,3,4,5,6,7,8,9,10,11,12,13,14,15,16,17,18,19,20,21,22,23,24]$ \\
\hline
\end{tabular}
\end{table}

\section{Application to rendezvous and proximity operations} \label{app}
In this section, we formulate the problem of guiding a chaser spacecraft to a target. While formulating this problem, the disturbances due to atmospheric drag, gravity gradient, and solar radiation pressure are neglected. Moreover, no external forces are considered other than the thrust force.

\subsection{Translational Dynamics: Tschauner-Hempel Model}

\begin{figure}[htbp]
	\centering{\includegraphics[trim={2cm 6cm 8cm 0},clip,width=0.8\columnwidth]{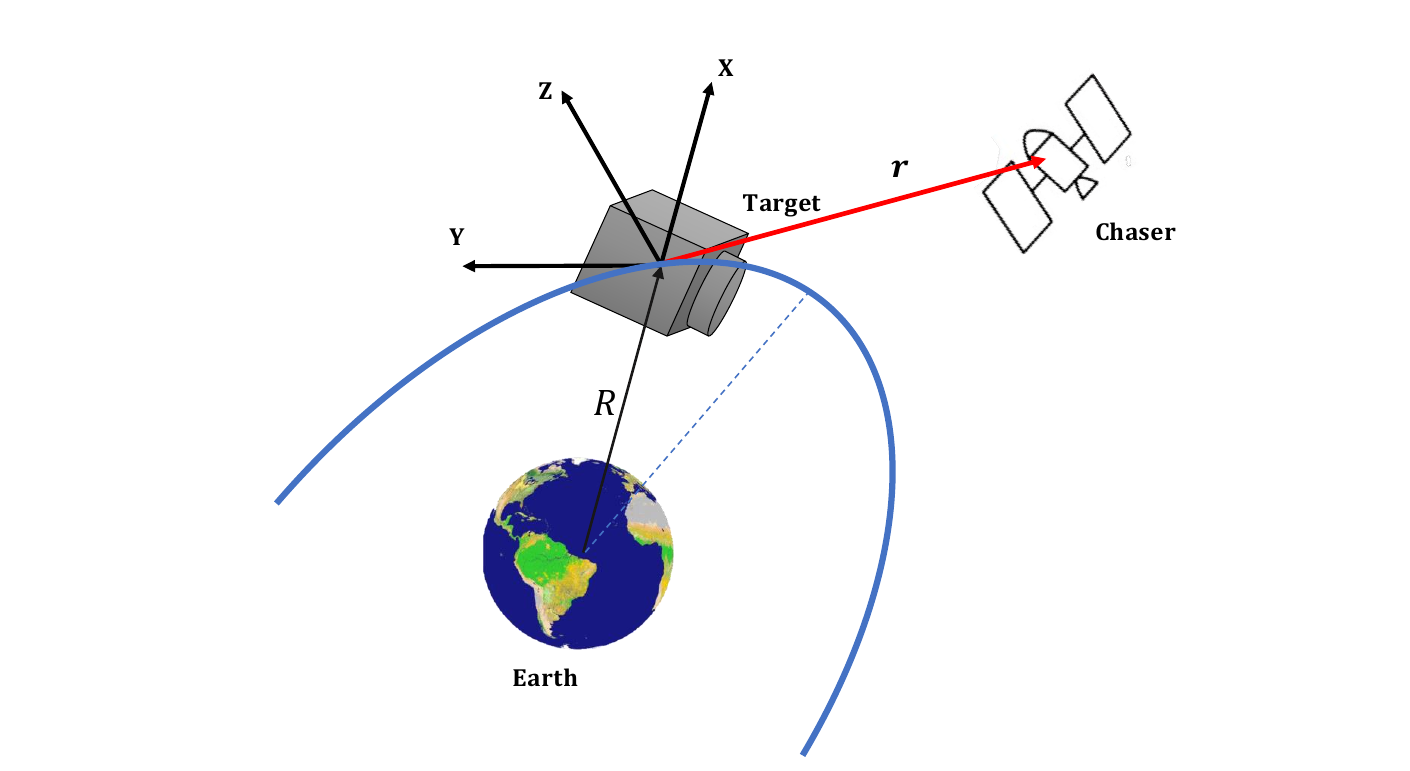}}
	\caption{LVLH frame of reference for a target.}
	\label{lvlh}
\end{figure}

The translational relative dynamics are modeled using the Tschauner-Hempel equations, which describe the motion of a chaser spacecraft with respect to a target on an elliptical orbit. These equations are formulated in the local-vertical-local-horizontal (LVLH) frame fixed to the target as shown in Fig.~\ref{lvlh}, where the $x$-axis points along-track, the $y$-axis lies in the cross-track direction, and the $z$-axis is normal to the orbital plane. 

The nonlinear equations of motion are given by:
\begin{equation} \label{eq:23}
	\begin{array}{c}
		\ddot{x}-2 \dot{\theta} \dot{y}-\ddot{\theta} y-\dot{\theta}^2 x-2 \mu x \left({\frac{\dot{\theta}}{h}}\right)^{3/2} = a_x = u_x \\\\
		\ddot{y}+2 \dot{\theta} \dot{x}-\ddot{\theta} x+\dot{\theta}^2 y+\mu y \left({\frac{\dot{\theta}}{h}}\right)^{3/2} = a_y = u_y \\\\
		\ddot{z}+ \mu z \left({\frac{\dot{\theta}}{h}}\right)^{3/2} = a_z = u_z
	\end{array}
\end{equation}
where $\theta$ is the true anomaly, $h$ is the specific angular momentum of the target orbit, $\mu$ is the Earth's gravitational constant, and $\{u_x, u_y, u_z\}$ are the control accelerations applied by the chaser in the LVLH frame.

Linearizing Eq.~\eqref{eq:23} leads to a time-varying state-space system of the form:
\begin{equation} \label{eq:24}
	\dot{\boldsymbol{x}}_t = \boldsymbol{A}_t \boldsymbol{x}_t + \boldsymbol{B}_t \boldsymbol{u}_t, \quad \boldsymbol{y}_t = \boldsymbol{C}_t \boldsymbol{x}_t
\end{equation}
where the state, input, and output vectors are defined as $\boldsymbol{x}_t = [x, y, z, \dot{x}, \dot{y}, \dot{z}]^T$, $\boldsymbol{u}_t = [u_x, u_y, u_z]^T$, and $\boldsymbol{y}_t = [x, y, z]^T$. The system matrices are given by:
\begin{equation}
	\boldsymbol{A}_t =
	\begin{bmatrix}
		0 & 0 & 0 & 1 & 0 & 0 \\
		0 & 0 & 0 & 0 & 1 & 0 \\
		0 & 0 & 0 & 0 & 0 & 1 \\
		\frac{3 + e \cos \theta}{1 + e \cos \theta} \dot{\theta}^2 & \ddot{\theta} & 0 & 0 & 2 \dot{\theta} & 0 \\
		-\ddot{\theta} & \frac{e \cos \theta}{1 + e \cos \theta} \dot{\theta}^2 & 0 & -2 \dot{\theta} & 0 & 0 \\
		0 & 0 & -\frac{1}{1 + e \cos \theta} \dot{\theta}^2 & 0 & 0 & 0
	\end{bmatrix}
\end{equation}
\begin{equation}
	\boldsymbol{B}_t =
	\begin{bmatrix}
		0 & 0 & 0 \\
		0 & 0 & 0 \\
		0 & 0 & 0 \\
		1 & 0 & 0 \\
		0 & 1 & 0 \\
		0 & 0 & 1
	\end{bmatrix}, \quad
	\boldsymbol{C}_t = 
	\begin{bmatrix}
		1 & 0 & 0 & 0 & 0 & 0 \\
		0 & 1 & 0 & 0 & 0 & 0 \\
		0 & 0 & 1 & 0 & 0 & 0
	\end{bmatrix}
\end{equation}

The continuous-time system is discretized with sampling time $T_s$ using matrix exponential and integral methods:
\begin{equation}
	\boldsymbol{x}_{t,k+1} = \boldsymbol{A}_{t,k} \boldsymbol{x}_{t,k} + \boldsymbol{B}_{t,k} \boldsymbol{u}_{t,k}
\end{equation}
\begin{equation}
	\boldsymbol{A}_{t,k} = e^{\boldsymbol{A}_t T_s}, \quad
	\boldsymbol{B}_{t,k} = \left( \int_0^{T_s} e^{\boldsymbol{A}_t \tau} d\tau \right) \boldsymbol{B}_t
\end{equation}
The matrices $\boldsymbol{A}_{t,k}$ and $\boldsymbol{B}_{t,k}$ are updated at each control step. 

\subsection{Rotational Dynamics: Relative Attitude Motion}

For precise orientation control during rendezvous and proximity operation (RPO), the rotational motion of the chaser is modeled relative to the target spacecraft. The relative attitude is described using a scalar-first quaternion $\boldsymbol{q}_{\text{rel}} = [q_0, q_1, q_2, q_3]^T$, defined as the rotation from the target to the chaser:
\begin{equation}
    \boldsymbol{q}_{\text{rel}} = \boldsymbol{q}_T^{-1} \otimes \boldsymbol{q}_C
\end{equation}
where $\boldsymbol{q}_T$ and $\boldsymbol{q}_C$ are the target and chaser quaternions, and $\otimes$ denotes quaternion multiplication.

The relative angular velocity in the target frame is:
\begin{equation}
    \boldsymbol{\omega}_{\text{rel}} = \boldsymbol{\omega}_C - \boldsymbol{R}_{\text{rel}} \boldsymbol{\omega}_T
\end{equation}
where $\boldsymbol{R}_{\text{rel}}$ is the rotation matrix corresponding to $\boldsymbol{q}_{\text{rel}}$, and $\boldsymbol{\omega}_C$, $\boldsymbol{\omega}_T$ are the chaser and target angular velocities, respectively.

The relative quaternion kinematics are:
\begin{equation}
    \dot{\boldsymbol{q}}_{\text{rel}} = \frac{1}{2} {\bm \Omega}(\boldsymbol{\omega}_{\text{rel}}) \boldsymbol{q}_{\text{rel}}
\end{equation}
with the skew-symmetric matrix ${\bm \Omega}(\boldsymbol{\omega})$ given by:
\begin{equation}
    {\bm \Omega}(\boldsymbol{\omega}) =
    \begin{bmatrix}
        0 & -\omega_x & -\omega_y & -\omega_z \\
        \omega_x & 0 & \omega_z & -\omega_y \\
        \omega_y & -\omega_z & 0 & \omega_x \\
        \omega_z & \omega_y & -\omega_x & 0
    \end{bmatrix}
\end{equation}

Assuming only the chaser applies torques, the relative angular velocity dynamics follow:
\begin{equation}
    \dot{\boldsymbol{\omega}}_{\text{rel}} = \boldsymbol{I}^{-1} \left( \boldsymbol{\tau} - \boldsymbol{\omega}_{\text{rel}} \times (\boldsymbol{I} \boldsymbol{\omega}_{\text{rel}}) \right)
\end{equation}
where $\boldsymbol{I}$ is the chaser's inertia matrix and $\boldsymbol{\tau}$ is the control torque in the body frame.

The relative rotational state vector is $\boldsymbol{x}_a = [q_0, q_1, q_2, q_3, \omega_x, \omega_y, \omega_z]^T$, and the control input is $\boldsymbol{u}_a = [\tau_x, \tau_y, \tau_z]^T$. Linearization around the current estimate yields:
\begin{equation}
    \dot{\boldsymbol{x}}_a = \boldsymbol{A}_a \boldsymbol{x}_a + \boldsymbol{B}_a \boldsymbol{u}_a
\end{equation}
where $\boldsymbol{A}_a$ and $\boldsymbol{B}_a$ are the Jacobian matrices evaluated at each control step.

\subsection{Integrated 6-DOF Dynamics and Control Framework}

Combining the Tschauner-Hempel translational model with the relative rotational dynamics results in a unified 6-DOF system for RPO:
\begin{equation}
    \dot{\boldsymbol{x}} =
    \begin{bmatrix}
        \boldsymbol{A}_t & \boldsymbol{0} \\
        \boldsymbol{0} & \boldsymbol{A}_a
    \end{bmatrix}
    \boldsymbol{x} +
    \begin{bmatrix}
        \boldsymbol{B}_t & \boldsymbol{0} \\
        \boldsymbol{0} & \boldsymbol{B}_a
    \end{bmatrix}
    \boldsymbol{u}
    =
    {\bm A}_c {\bm x} + {\bm B}_c {\bm u}
\end{equation}
where the full state vector is $\boldsymbol{x} = [\boldsymbol{x}_t^T, \boldsymbol{x}_a^T]^T$ and the control input is $\boldsymbol{u} = [\boldsymbol{u}_t^T, \boldsymbol{u}_a^T]^T$.

Here, $\boldsymbol{x}_t$ represents the relative position and velocity of the chaser with respect to the target in the LVLH frame, while $\boldsymbol{x}_a$ describes the relative attitude (quaternion) and angular velocity in the target's body frame. The combined model is discretized at each control step using matrix exponential and integral methods. As our control strategy, model predictive control (MPC) is introduced in the next section.

\section{Model Predictive Control Formulation for 6-DOF Rendezvous and Docking}\label{sec:mpc}
The basic idea of the MPC is to numerically optimize a model to obtain a control input sequence that minimizes a cost function over a finite horizon. The control actions are periodically re-computed at each sampling instant with the current state estimate as an initial condition, thereby providing a feedback action.

\subsection{Prediction Model}

At each sampling instant $k$, we linearize the full nonlinear relative dynamics of the chaser spacecraft about the current operating point $\left(\mathbf{x}_k, \mathbf{f}_{k-1}\right)$. The state vector is defined as
\begin{equation} \label{eq:xk_def}
{\bm x}_k = 
\begin{bmatrix}
    \mathbf{r}_k \\
    \dot{\mathbf{r}}_k \\
    \mathbf{q}_{\text{rel},k} \\
    \boldsymbol{\omega}_{\text{rel},k}
\end{bmatrix}
=
\begin{bmatrix}
    x, \; y, \; z, \; \dot{x}, \; \dot{y}, \; \dot{z}, \; q_0, \; q_1, \; q_2, \; q_3, \; \omega_x, \; \omega_y, \; \omega_z
\end{bmatrix}^{\mathrm{T}} \in \mathbb{R}^{13},
\end{equation}

where $(x, y, z)$ denotes relative position in the LVLH frame, ${\bm q}_{\text{rel},k} = [q_0, q_1, q_2, q_3]^T$ is the unit quaternion representing chaser attitude with respect to LVLH, and $\boldsymbol{\omega}_{\text{rel},k}$ is the body angular velocity.

The control input is the vector of individual thruster forces:
\begin{equation}
    {\bm u}_{f_k} = \begin{bmatrix}
        f_{1,k}\;, & \cdots\;, & f_{N,k}
    \end{bmatrix}^T \in \mathbb{R}^N,
\end{equation}
mapped to force and torque by
\begin{equation}
    \begin{bmatrix}
        {\bm f}_k \\ \boldsymbol{\tau}_k
    \end{bmatrix}
    = {\bm B}_{\text{alloc}} {\bm u}_{f_k},
\end{equation}
where ${\bm B}_{\text{alloc}} \in \mathbb{R}^{6 \times N}$ denotes the thruster allocation matrix, defined as the active-component subset of $\mathcal{\bm H}$ for consistency of notation, and is computed from the configuration geometry identified through the feasible configuration search.

The discrete-time linear time-varying dynamics are given by:
\begin{equation}
    {\bm x}_{k+1} = {\bm A}_k {\bm x}_k + {\bm B}_k {\bm u}_{f_k},
\end{equation}
where ${\bm A}_k = e^{{\bm A}_c(\theta_k) T_s}$ and ${\bm B}_k = \left( \int_0^{T_s} e^{{\bm A}_c(\theta_k) \tau} d\tau \right) {\bm B}_c {\bm B}_{\text{alloc}}$. The matrices depend on the true anomaly $\theta_k$, quaternion, and angular velocity, and are updated at each time step.

\subsection{Constraints}

\paragraph{Thruster bounds:} Each thruster has a unidirectional bounded output:
\begin{equation}
    f_{\min} \leq f_{i,k} \leq f_{\max}, \quad \forall i = 1, \ldots, N.
\end{equation}

\paragraph{Quaternion normalization:} A soft constraint is used to enforce unit norm:
\begin{equation}
    \varphi_k = \left( {\bm q}_k^T {\bm q}_k - 1 \right) \quad \Rightarrow \quad \ell_{\text{quat}} = \rho \varphi_k^2.
\end{equation}

\paragraph{Terminal docking constraint:} The terminal state is constrained to a neighborhood:
\begin{equation}
    {\bm x}_{k+N_{mpc}} \in \mathcal{E}_f = \left\{ {\bm x} ~\middle|~ 
        \| {\bm r} \| \leq r_{\text{tol}}, ~
        \| \dot{{\bm r}} \| \leq v_{\text{tol}}, ~
        \angle({\bm q}, {\bm q}_{\text{dock}}) \leq \alpha_{\text{tol}} \right\}.
\end{equation}

\subsection{Cost Function}

Define the deviation from the reference:
\begin{equation}
    \tilde{{\bm x}}_{k+i} = {\bm x}_{k+i|k} - {\bm x}^{\text{ref}}_{k+i}, \quad
    \tilde{{\bm f}}_{k+i} = {\bm f}_{k+i|k} - {\bm f}^{\text{ref}}_{k+i}.
\end{equation}

The cost at each stage includes state tracking, thrust minimization, and slew penalties:
\begin{equation}
    \ell_{k+i} = \tilde{{\bm x}}_{k+i}^T {\bm Q} \tilde{{\bm x}}_{k+i}
    + \tilde{{\bm f}}_{k+i}^T {\bm R} \tilde{{\bm f}}_{k+i}
    + \Delta {\bm f}_{k+i}^T {\bm R}_{\dot{f}} \Delta {\bm f}_{k+i}, \quad
    \Delta {\bm f}_{k+i} = {\bm f}_{k+i|k} - {\bm f}_{k+i-1|k},
\end{equation}
with terminal cost $\ell_f = \tilde{{\bm x}}_{k+N_{mpc}}^T {\bm P} \tilde{{\bm x}}_{k+N_{mpc}}$. The weight matrices are phase-dependent:
\begin{equation}
    ({\bm Q}, {\bm R}, {\bm R}_{\dot{f}}) = 
    \begin{cases}
        ({\bm Q}_1, {\bm R}_1, {\bm R}_{\dot{f},1}), & \text{Approach phase} \\
        ({\bm Q}_2, {\bm R}_2, {\bm R}_{\dot{f},2}), & \text{Final docking}
    \end{cases}.
\end{equation}

\subsection{MPC Optimization Problem}

The finite-horizon MPC problem at each time step is:
\begin{equation}
\begin{aligned}
    \min_{\{{\bm f}_{k+i|k}\}_{i=0}^{N_{mpc}-1}} \quad &
    \sum_{i=0}^{N_{mpc}-1} \left[
        \tilde{{\bm x}}_{k+i}^T {\bm Q} \tilde{{\bm x}}_{k+i}
        + \tilde{{\bm f}}_{k+i}^T {\bm R} \tilde{{\bm f}}_{k+i}
        + \Delta {\bm f}_{k+i}^T {\bm R}_{\dot{f}} \Delta {\bm f}_{k+i}
    \right]
    + \tilde{{\bm x}}_{k+N_{mpc}}^T {\bm P} \tilde{{\bm x}}_{k+N_{mpc}}
    + \rho \varphi_{k+N_{mpc}}^2 \\
    \text{s.t.} \quad & {\bm x}_{k|k} = {\bm x}_k, \\
    & {\bm x}_{k+i+1|k} = {\bm A}_{k+i} {\bm x}_{k+i|k} + {\bm B}_{k+i} {\bm f}_{k+i|k}, \quad i = 0,\dots,N_{mpc}-1, \\
    & f_{\min} \leq f_{i,k+j|k} \leq f_{\max}, \quad i = 1,\dots,n,~ j = 0,\dots,N_{mpc}-1, \\
    & {\bm x}_{k+N_{mpc}|k} \in \mathcal{E}_f \quad \text{(if final phase)}.
\end{aligned}
\end{equation}

This problem is a convex Quadratic Program (QP) due to linear dynamics and quadratic cost, and is solved efficiently at each step.

\section{From Configuration to Control: System Workflow}
Before presenting the numerical results, we briefly summarise the complete workflow that links the \emph{offline} thruster–layout search with the \emph{online} MPC docking controller. First, an exhaustive offline sweep is carried out to identify a feasible thruster configuration for the cubic satellite of side–length~$L$. All $24$ candidate mounting locations on the cube’s faces are considered, and every combination of exactly $N$ thrusters is paired with every admissible orientation chosen from the discretised angle grid~$\{\theta_j,\phi_j\}$. For each (layout, orientation) pair, the corresponding $6\times N$ control allocation matrix~$\mathcal{\bm H}_\ell$ is assembled; if its rank is less than six the pair is immediately discarded because full 6-DOF authority is impossible. Otherwise the pair is tested against a set of twelve unit wrenches, $\pm$\,unit forces along the body–fixed $\bm X$, $\bm Y$, $\bm Z$ axes and $\pm$\,unit torques about those axes. Each wrench is reproduced by solving a \textit{non–negative least squares} problem, and the residual must fall below the prescribed tolerance~$\epsilon_{\mathrm{NNLS}}$. A pair that satisfies this criterion for \emph{all} twelve wrenches is added to the viable–configuration set~$\mathcal{V}$. When the sweep finishes, the configuration that achieves the minimum cumulative thrust across the twelve unit–wrench tests is selected and its allocation sub–matrix~${\bm B}_{\text{alloc}}$ (i.e. $\mathcal{\bm H}_\ell$) is stored for flight use.

The \emph{online} phase employs this fixed allocation together with a linear–time–varying MPC scheme. At each control step~$k$ the current state vector~${\bm x}_{k}$ is measured, and the time–varying linearised dynamics matrices ${\bm A}_{k},{\bm B}_{k}$ are refreshed using the latest relative–orbit parameters. With these matrices the MPC optimiser constructs prediction models over the horizon length~$N_{mpc}$ and solves a quadratic programme that minimises a weighted sum of state–tracking error, control effort, and thrust–rate penalties. The first element of the optimal thrust sequence ${\bm f}^{\star}_{k:k+N_{mpc}-1}$ is applied, the full nonlinear spacecraft dynamics are propagated over the sampling interval~$T_{s}$, and the process repeats. When intermediate waypoint tolerances are met, the cost weights are smoothly switched to a tighter terminal set $({\bm Q}_{2},{\bm R}_{2},{\bm R}_{\dot f,2})$ to refine the final–phase approach. The loop terminates once the docking criteria (position, attitude, and relative–velocity bands) are simultaneously satisfied, guaranteeing closed–loop 6–DOF control authority throughout the rendezvous and contact phase.

\section{Simulation Result}

\begin{table}[htb!]
\small
\centering
\caption{Simulation Parameters for Rendezvous and Docking}
\label{tab:sim_params}
\begin{tabular}{llc}
\hline
\textbf{Parameter} & \textbf{Description} & \textbf{Value} \\
\hline
$L$ & Chaser side length & $0.5~\text{m}$ \\
$m$ & Chaser mass & $20~\text{kg}$ \\
$I$ & Moment of inertia (diagonal) & $\frac{1}{6}mL^2~\text{diag}(1,1,1)$ \\
$\mu$ & Earth gravitational parameter & $3.986 \times 10^{14}~\text{m}^3/\text{s}^2$ \\
$a_e$ & Target semi-major axis & $12,000~\text{km}$ \\
$e$ & Target orbit eccentricity & $0.1$ \\
$h$ & Specific angular momentum & $\sqrt{\mu a_e (1 - e^2)}$ \\
$\delta t$ & Time step & $0.1~\text{s}$ \\
$T_f$ & Final simulation time & $400~\text{s}$ \\
$N_{mpc}$ & MPC prediction horizon & $10$ steps \\
$n_{\text{state}}$ & State dimension & $13$ \\
$N$ & Number of thrusters & Varies (e.g., $8$–$24$) \\

$f_{\min}$ & Minimum thruster force & $0~\text{N}$ \\
$f_{\max}$ & Maximum thruster force & $0.05~\text{N}$ \\
$Q_1$ & State weight (Phase 1) & $100\times\text{diag}(8,8,8,\,8,8,8,\,5,5,5,5,\,5,5,5)$ \\
$R_1$ & Input weight (Phase 1) & $500 \cdot \bm{I}$ \\
$R_{\dot{f},1}$ & Input rate weight (Phase 1) & $1000 \cdot \bm{I}$ \\
$Q_2$ & State weight (Phase 2) & $10^4\times\text{diag}(0.9,0.9,0.9,\,5,5,5,\,10,10,10,10,\,1,1,1)$ \\
$R_2$ & Input weight (Phase 2) & $5 \times 10^4 \cdot \bm{I}$ \\
$R_{\dot{f},2}$ & Input rate weight (Phase 2) & $5 \times 10^4 \cdot \bm{I}$ \\
\hline
\end{tabular}
\end{table}

\begin{figure}[htb!]
    \centering
    \begin{subfigure}{0.3\textwidth}
        \centering
        \includegraphics[trim={2cm, 1cm, 0, 1cm}, width=1.0\textwidth]{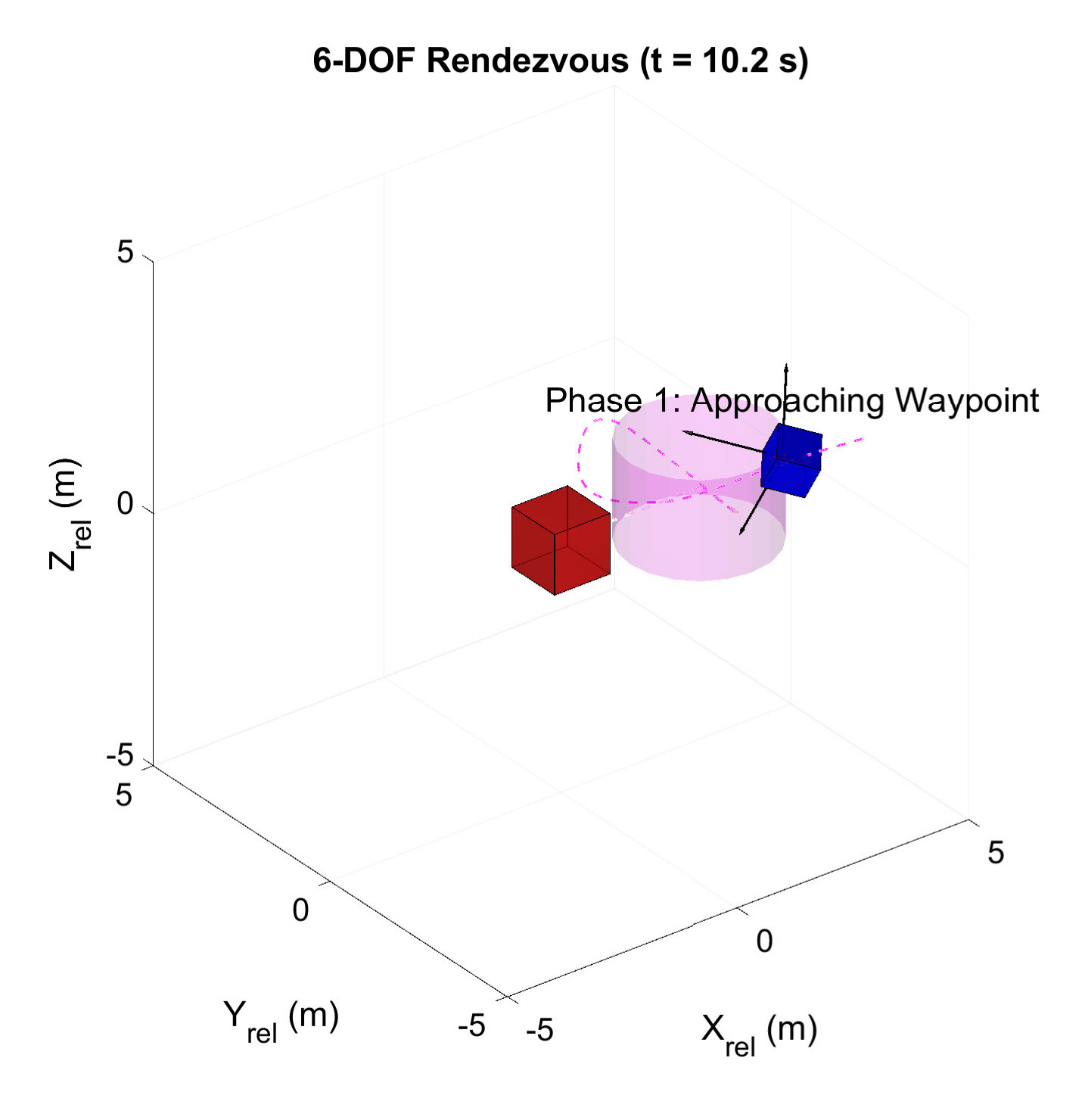}
        \caption{Phase 1: Waypoint approach}
        \label{fig:config7_frame1}
    \end{subfigure}
    \begin{subfigure}{0.3\textwidth}
        \centering
        \includegraphics[trim={1cm, 1cm, 1cm, 1cm}, width=1.0\textwidth]{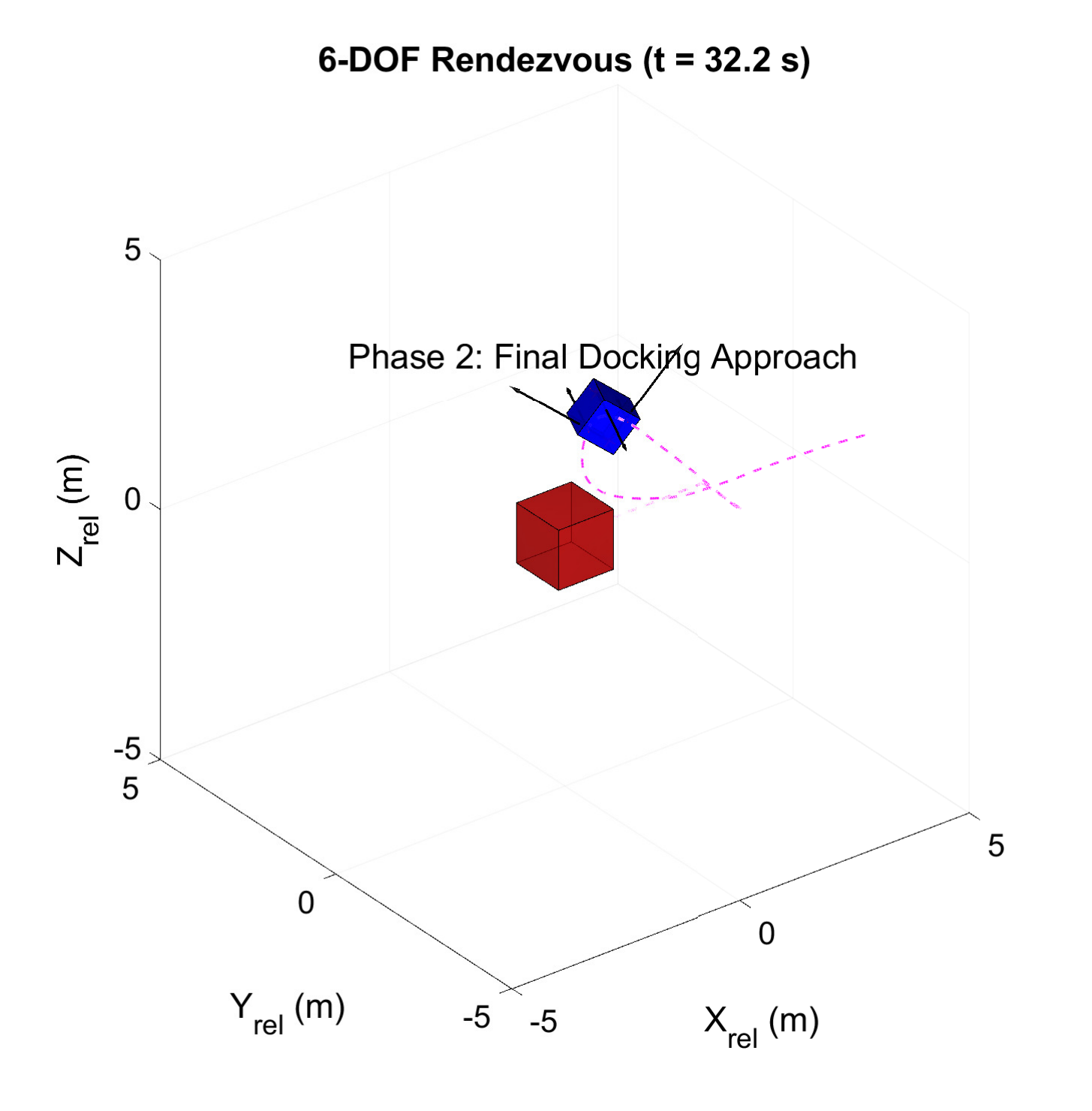}
        \caption{Phase 2: Docking approach}
        \label{fig:config7_frame2}
    \end{subfigure}
    \begin{subfigure}{0.3\textwidth}
        \centering
        \includegraphics[trim={0, 1cm, 2cm, 1cm}, width=1.0\textwidth]{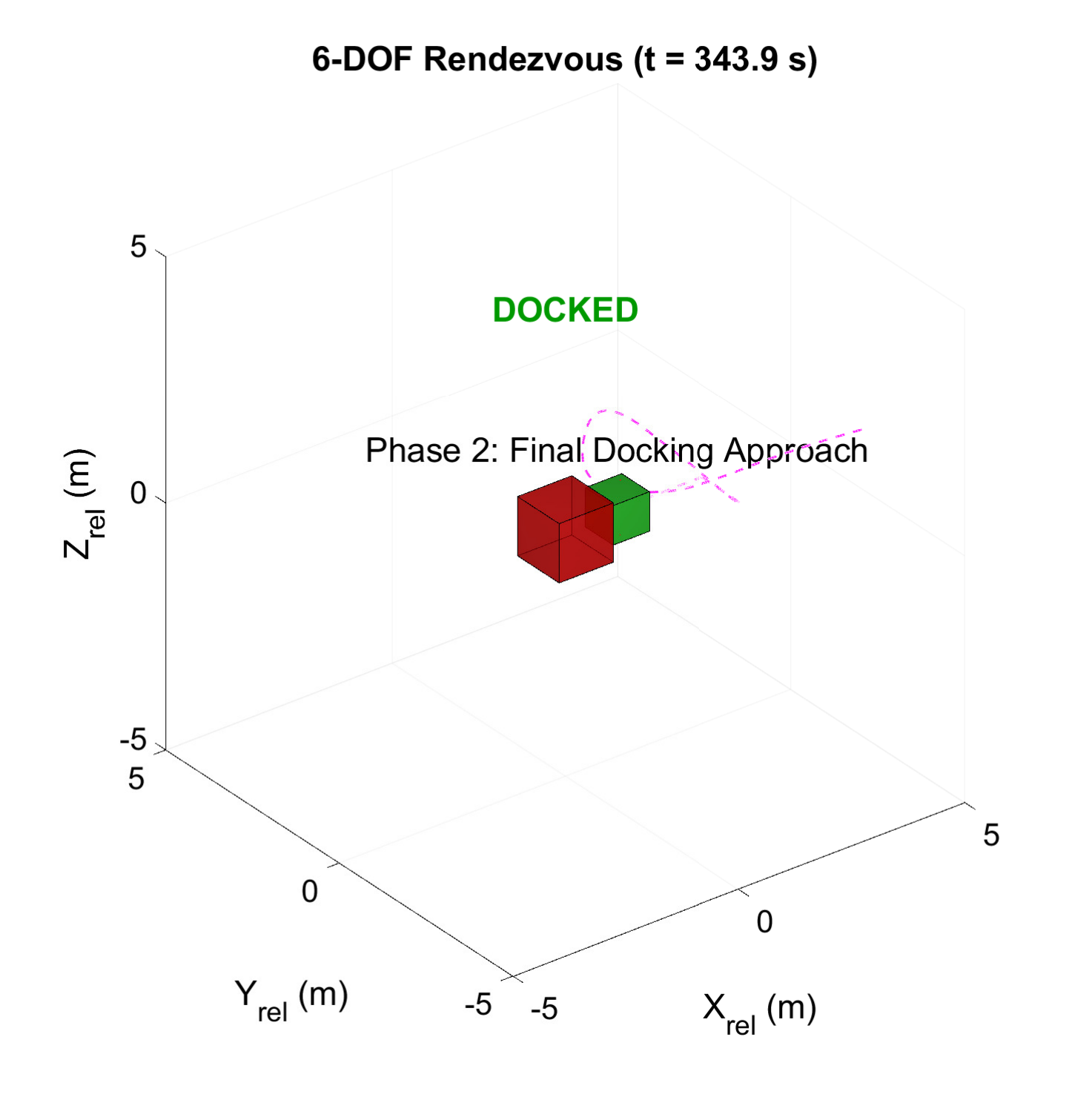}
        \caption{Docking}
        \label{fig:config7_frame4}
    \end{subfigure}
    \caption{Simulation snapshots by phases}
    \label{fig:sim_snapshots}
\end{figure}

\begin{figure}[htb!]
    \centering
    \begin{subfigure}{0.82\textwidth}
        \centering
        \includegraphics[trim={4cm, 0, 4cm, 0},width=1.0\textwidth]{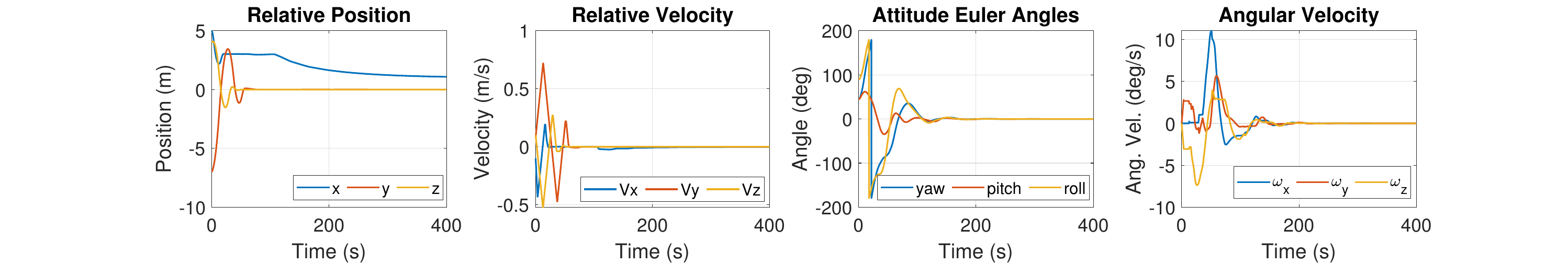}
        \caption{\(N=7\) optimal config. state profile}
        \label{fig:config7_state_profile}
    \end{subfigure}
    \begin{subfigure}{0.82\textwidth}
        \centering
        \includegraphics[trim={4cm, 0, 4cm, 0},width=1.0\textwidth]{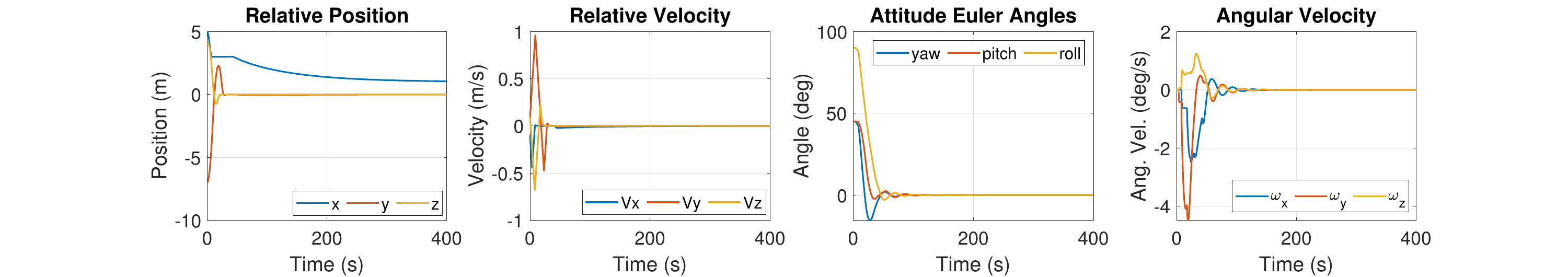}
        \caption{\(N=12\) optimal config. state profile}
        \label{fig:config12_state_profile}
    \end{subfigure}
    \begin{subfigure}{0.82\textwidth}
        \centering
        \includegraphics[trim={4cm, 0, 4cm, 0},width=1.0\textwidth]{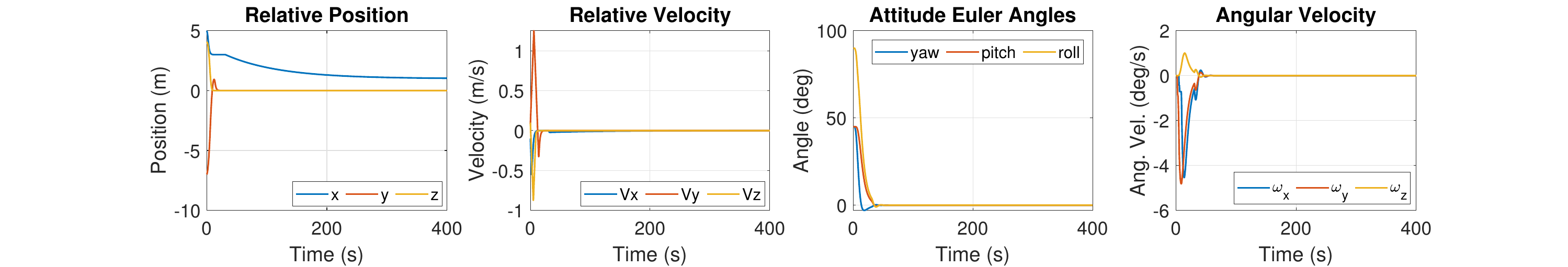}
        \caption{\(N=24\) optimal config. state profile}
        \label{fig:config24_state_profile}
    \end{subfigure}
    \caption{State profile comparison of the least thruster count configuration (\(N=7\)), the optimal thruster count configuration (\(N=12\)), and the initial configuration (\(N=24\))}
    \label{fig:state_profile}
\end{figure}

The rendezvous-docking simulation is conducted with the parameters shown in Table~\ref{tab:sim_params}, and their results are compared by arbitrarily selecting one of the optimal configurations from each of the configurations from \(N = 7\) to \(24\). The results of the simulations are compared based on the time it takes for a satellite to complete the mission and the total impulse required for the entire mission, as shown in Table~\ref{tab:sim_outcome}. Moreover, given these results and the previous results shown in Table~\ref{tab:outcome}, the behaviors and thruster firing profiles of configurations with the characteristic number of thrusters, that is 7, 12, and 24, are shown in Fig.~\ref{fig:state_profile}. Snapshots of the simulation are provided in Fig.~\ref{fig:sim_snapshots}.

Table~\ref{tab:sim_outcome} summarizes the performance of thruster configurations ranging from 6 to 24 thrusters by docking time and total impulse. The 7-thruster configuration, the minimum number of thrusters for 6-DOF control, achieves docking in 333.5 seconds with a total impulse of 12.9784 N-s. As the thruster count increases, configurations with 12 or more thrusters markedly reduce both docking time and total impulse; the 12-thruster case achieves docking in 291.2 s with 6.6160 N·s, indicating an optimal balance between control authority and efficiency. Beyond 12 thrusters, both docking time and total impulse remain less than those of configurations with 11 thrusters or fewer. This suggests that the 12-thruster configuration could offer an optimal balance between control authority and thrust efficiency, as additional thrusters beyond 12. These results also reinforcing the optimality of the 12-thruster configuration for efficient 6-DOF control in rendezvous and proximity operations.

\begin{table}[htb!]
\small
\centering
\caption{Summary of the simulation results}
\label{tab:sim_outcome}
\begin{tabular}{ccc}
\hline
\textbf{Number of Thrusters} & \textbf{Time to Dock (s)} & \textbf{Total Impulse (N-s)} \\
\hline
6  & -- & -- \\
7  & 343.9 & 12.9784 \\
8  & 331.5 & 10.2028 \\
9  & 306.5 & 7.9689 \\
10 & 315.0 & 7.7845 \\
11 & 337.0 & 10.1066 \\
12 & 291.2 & 6.6219 \\
13 & 292.5 & 6.9309 \\
14 & 291.5 & 6.9564 \\
15 & 285.5 & 6.2985 \\
16 & 287.5 & 5.8667 \\
17 & 277.0 & 5.8803 \\
18 & 267.5 & 5.6907 \\
19 & 275.5 & 7.2985 \\
20 & 267.5 & 6.6163 \\
21 & 263.5 & 7.1193 \\
22 & 267.7 & 7.2954 \\
23 & 270.6 & 7.6253 \\
24 & 264.6 & 6.6160 \\
\hline
\end{tabular}
\end{table}

\begin{figure}[htb!]
    \centering
    \begin{subfigure}{0.82\textwidth}
        \centering
        \includegraphics[trim={5cm, 0, 5cm, 0},width=1.0\textwidth]{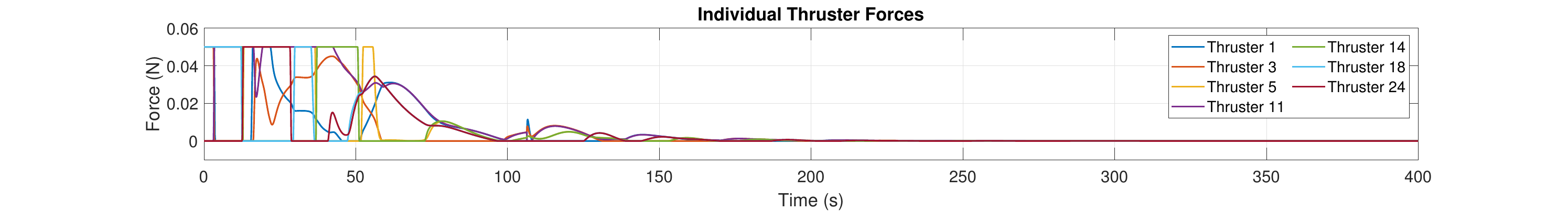}
        \caption{\(N=7\) optimal config. thrust profile}
        \label{fig:config7_thrust_profile}
    \end{subfigure}
    \begin{subfigure}{0.82\textwidth}
        \centering
        \includegraphics[trim={5cm, 0, 5cm, 0},width=1.0\textwidth]{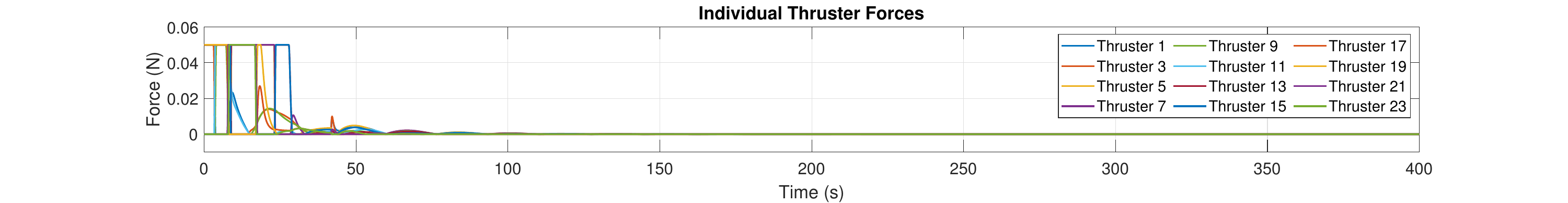}
        \caption{\(N=12\) optimal config. thrust profile}
        \label{fig:config12_thrust_profile}
    \end{subfigure}
    \begin{subfigure}{0.82\textwidth}
        \centering
        \includegraphics[trim={5cm, 0, 5cm, 0},width=1.0\textwidth]{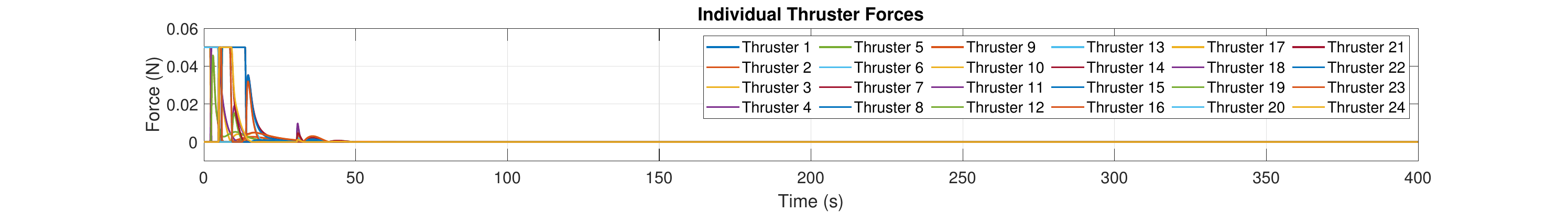}
        \caption{\(N=24\) optimal config. thrust profile}
        \label{fig:config24_state_profile}
    \end{subfigure}
    \caption{Thrust profile comparison of the least thruster count configuration (\(N=7\)), the optimal thruster count configuration (\(N=12\)), and the initial configuration (\(N=24\))}
    \label{fig:thrust_profile}
\end{figure}

The thrust and state profiles for the 7-thruster, 12-thruster, and 24-thruster configurations, illustrated in Fig.~\ref{fig:state_profile} and~\ref{fig:thrust_profile}, showcase distinct performance characteristics during the mission. The state profiles (Fig.~\ref{fig:state_profile}) show that all configurations successfully completed the docking mission by achieving the desired position and orientation. However, the 7-thruster configuration exhibits a longer time for convergence and more oscillation, compared to the 12-thruster configuration and the 24-thruster configuration. This difference is due to the limited control authority of the 7-thruster setup. While full 6-DOF controllability is ensured, it requires more time to execute precise maneuvers. 

The thrust profiles in Fig.~\ref{fig:thrust_profile} further indicate that the 7-thruster configuration demands longer peak thrust levels and more frequent thruster firings to compensate for its reduced number of actuators, resulting in the higher total impulse. On the other hand, the 12-thruster and 24-thruster configurations demonstrate smoother and more distributed thrust profiles within a shorter period as a result of leveraging their additional thrusters to achieve the same control objectives. While the time of mission completion and total impulse may vary depending on the selection of the optimal configuration, mission definitions, and constraints, these observations combined with the fact that the minimum total thrust (\(f_{N,min}\)) stabilizes at 30 N for \(N \geq12\) illustrate the efficiency of the optimal 12-thruster configuration, which not only reduces docking time but also minimizes propellant consumption, making it the preferred choice for practical implementation in this rendezvous and proximity operation.

\begin{figure}[htb!]
    \centering
    \includegraphics[width=1.0\textwidth]{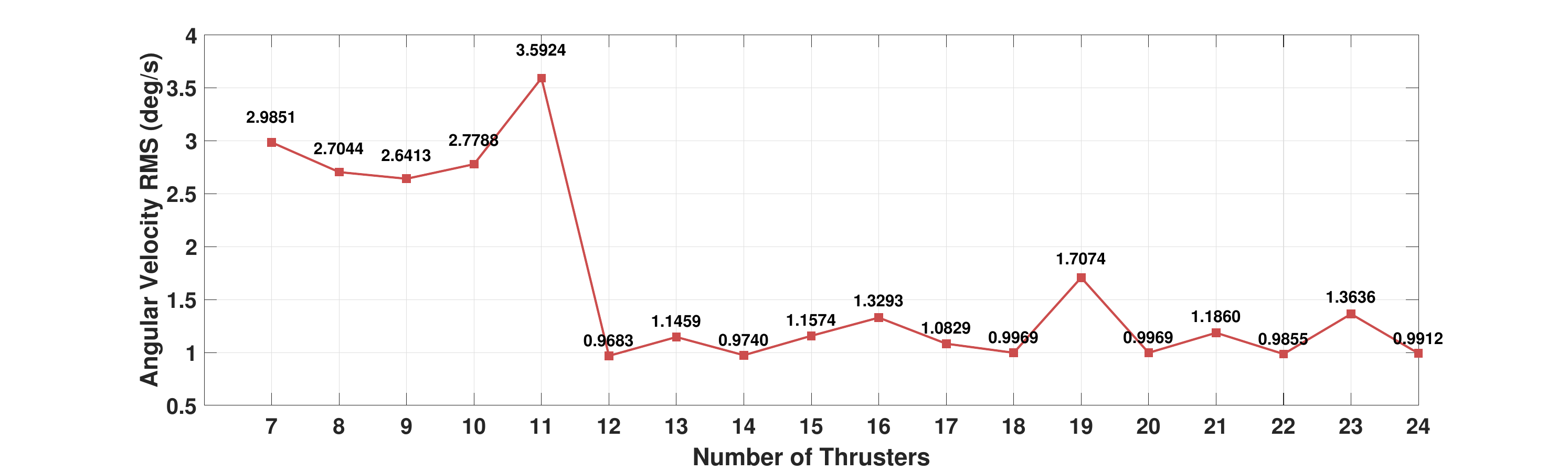}
    \caption{Root mean square of angular velocity for each thruster count}
    \label{fig:ang_vel_RMSE}
\end{figure}

Furthermore, Fig.~\ref{fig:ang_vel_RMSE} shows the root mean square (RMS) of the angular velocity for configurations ranging from 7 to 24 thrusters. As Fig.~\ref{fig:ang_vel_RMSE} shows, while angular velocity RMS stays higher for configurations with the thruster count of fewer than 12 thrusters, it significantly decreases beyond 12-thruster configurations. These patterns confirm that while higher thruster counts provide redundancy and smoother control, the 12-thruster optimal configuration could balance performance with minimal hardware, less complexity, and less fuel demands in missions requiring precise attitude stabilization.

\begin{figure}[htb!]
    \centering
    \begin{subfigure}{1.0\textwidth}
        \includegraphics[width=1.0\textwidth]{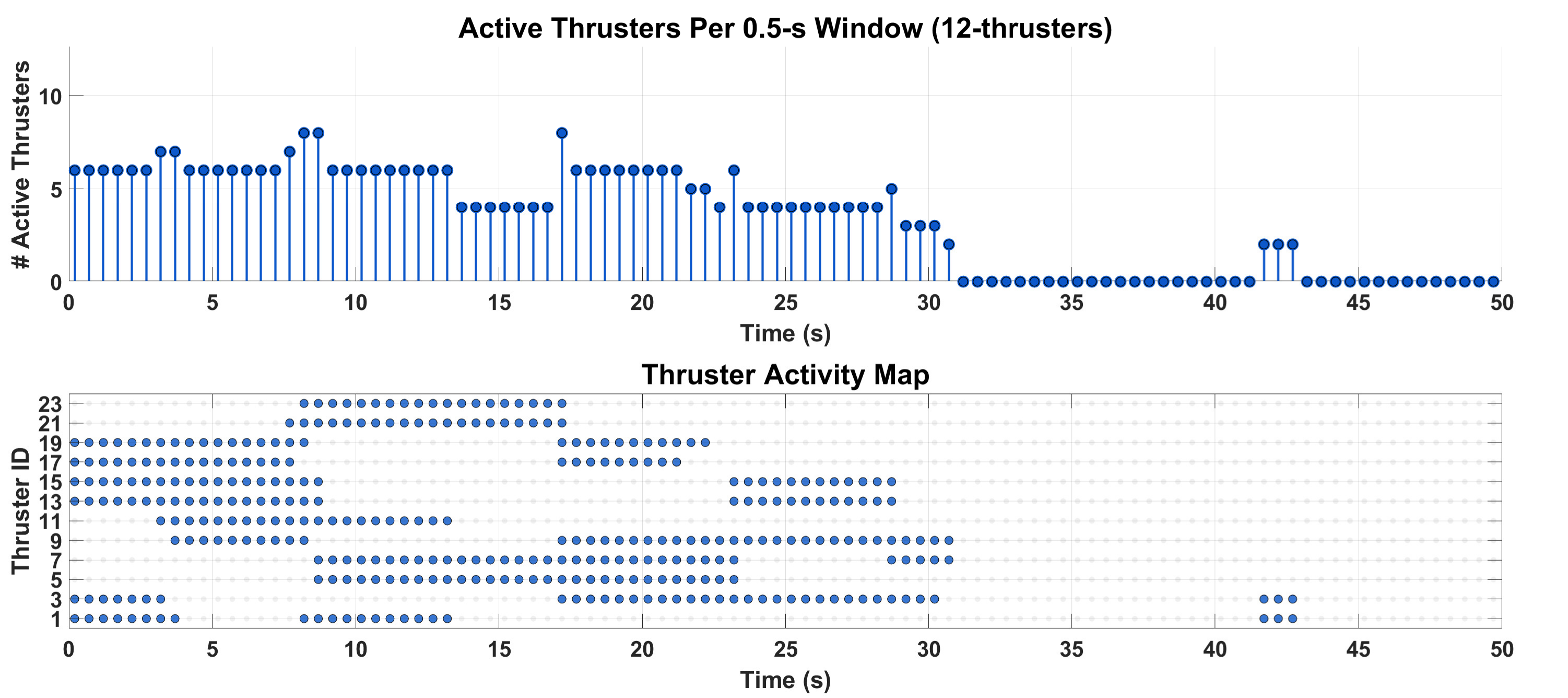}
        \caption{\(N=12\) active thruster and thruster activity map}
        \label{fig:config7_thrust_profile}
    \end{subfigure}
    \begin{subfigure}{1.0\textwidth}
        \includegraphics[width=1.0\textwidth]{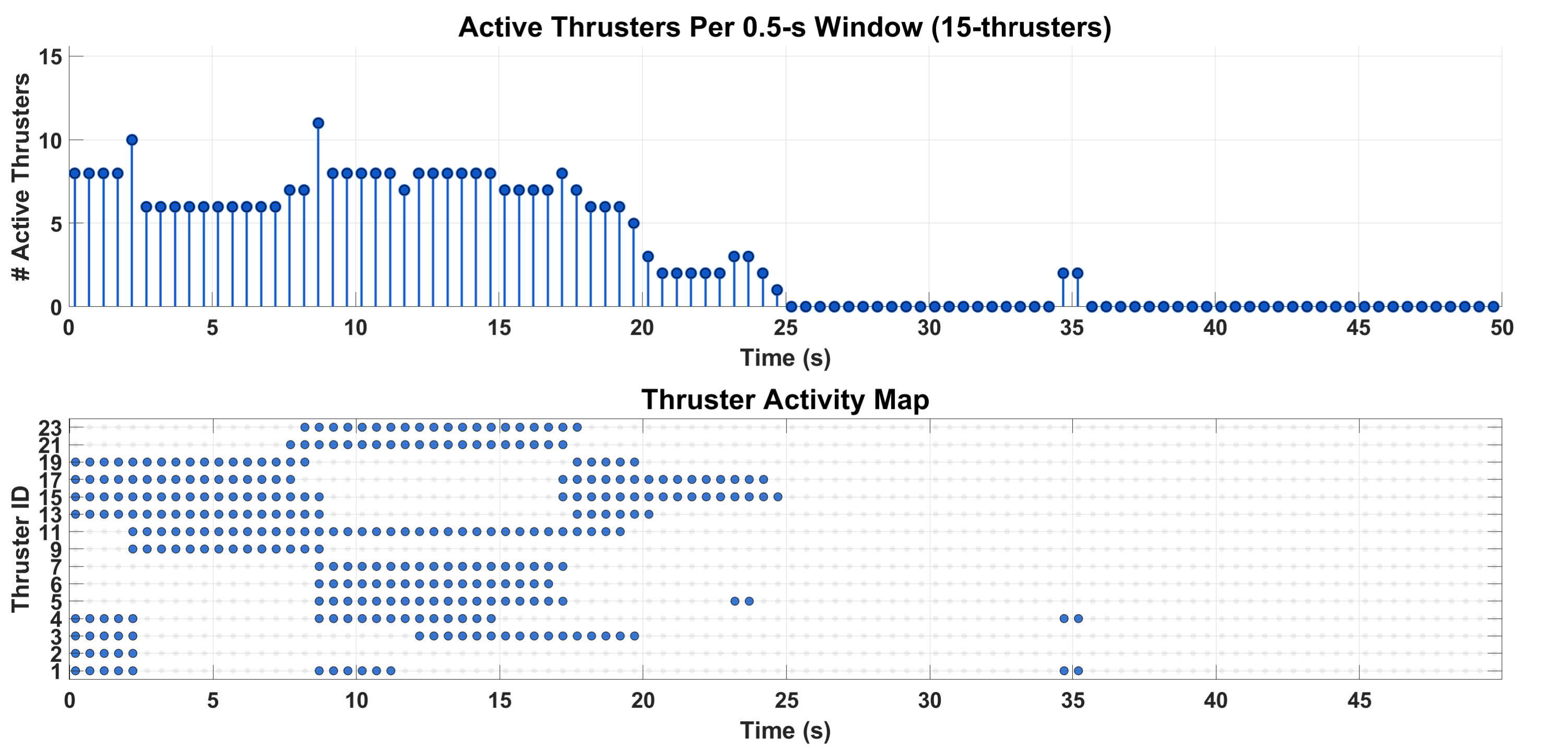}
        \caption{\(N=15\) active thruster and thruster activity ma}
        \label{fig:config12_thrust_profile}
    \end{subfigure}
    \caption{Active thruster and thruster activity map}
    \label{fig:active_thruster}
\end{figure}

Moreover, Fig.~\ref{fig:active_thruster} provides a detailed view of the active thrusters and their corresponding activity maps for the optimal 12-thruster configuration and a higher 15-thruster setup during the rendezvous-docking simulation. As it shows, although not all thrusters are activated at every instance, all of them are utilized throughout the simulation. In the 12-thruster case, the activity map shows a more balanced distribution of firing times across the thrusters, with no single actuator dominating the control effort. In contrast, the 15-thruster configuration exhibits clusters of underutilized thrusters. Although this setting could provide better fault tolerance in case of failures, this unbalance also indicate redundancies that may lead to inefficient fuel allocation and increased overall mass unnecessarily. This comparison showcases the trade-offs in thruster redundancy: the 12-thruster setup optimizes resource use for nominal operations, whereas the 15-thruster configuration offers robustness against incidents, such as thruster outages, at the cost of uneven load distribution. 



\section{Conclusion}
This study addresses the critical challenge of optimizing thruster configurations for achieving full six degrees of freedom control in small cubic satellites. By developing a heuristic algorithm that iteratively evaluates thruster placements and numbers using a nonnegative least squares approach, we identified the minimum and optimal number of thrusters required for complete 6-DOF controllability. The results confirm that a minimum of seven unidirectional thrusters is necessary to achieve 6-DOF control, aligning with theoretical predictions, which states \(n+1\) unidirectional actuators are required for \(n\)-DOF operation, while a 12-thruster configuration offers an optimal balance between control authority and thrust efficiency. The application of MPC framework, incorporating the Tschauner-Hempel equations and thruster allocation constraints, ensures precise guidance during rendezvous and docking phases. These findings provide valuable insights for designing efficient propulsion systems, reducing mass and complexity while maintaining robust 6-DOF control for missions such as space debris removal, satellite servicing, and other proximity operations. 

\section{Future Work}
As extensions of this study, the following further evaluations can be considered as future work. One of the future studies could consider dynamically changing thruster configurations. As mentioned previously, required thrust differs between missions even though the number of thrusters equipped to the satellite is consistent. Therefore, by installing 24 thrusters on a satellite, one can find an optimal configuration of an \(N\)-thruster setting to employ at each instant, ensuring optimality of the mission. Another extension could include a re-evaluation of the configurations with various shapes of satellites. Moreover, another future work could explore the impact of thruster gimbal capabilities and external disturbances on configuration performance to further enhance mission robustness.

\section{Acknowledgement}
The authors acknowledge the support received from Rogue Space Systems Corporation via award number FA864922P1201 for this work, which was conducted under an AFWERX STTR FX21S-TCS01 Phase I contract.

\bibliographystyle{AAS_publication}   
\bibliography{references}   

\end{document}